\def\cvprPaperID{4674}
\def\confName{CVPR}
\def\confYear{2023}

\def\paperTitle{HuManiFlow: Ancestor-Conditioned Normalising Flows on SO(3) Manifolds for Human Pose and Shape Distribution Estimation}

\def\authorBlock{
    Akash Sengupta\qquad
    Ignas Budvytis\qquad
    Roberto Cipolla \\
    University of Cambridge \\
    {\tt\small \{as2562, ib255, rc10001\}@cam.ac.uk}
}

\newif\ifreview 
\newif\ifarxiv \newcommand{\arxiv}{\arxivtrue}
\newif\ifcamera 
\newif\ifrebuttal 

\arxiv

\pdfoutput=1
\documentclass[10pt,twocolumn,letterpaper]{article}
\ifreview \usepackage[review]{cvpr} \fi
\ifarxiv \usepackage[pagenumbers]{cvpr} \fi
\ifrebuttal \usepackage[rebuttal]{cvpr} \fi
\ifcamera \usepackage{cvpr} \fi

\usepackage{graphicx}
\usepackage{amsmath}
\usepackage{amssymb}
\usepackage{booktabs}

\usepackage{times}
\usepackage{microtype}
\usepackage{epsfig}
\usepackage[table,xcdraw]{xcolor}
\usepackage{caption}
\usepackage{float}
\usepackage{placeins}
\usepackage{color, colortbl}
\usepackage{stfloats}
\usepackage{enumitem}
\usepackage{tabularx}
\usepackage{xstring}
\usepackage{multirow}
\usepackage{xspace}
\usepackage{url}
\usepackage{subcaption}
\usepackage{xcolor}
\usepackage[hang,flushmargin]{footmisc}

\ifcamera \usepackage[accsupp]{axessibility} \fi

% ...

\newcommand{\nbf}[1]{{\noindent \textbf{#1.}}}

\ifarxiv  \fi

\newcommand{\R}[1]{{%
    \textbf{%
        \ifstrequal{#1}{1}{\textcolor{red}{R#1}}{%
        \ifstrequal{#1}{2}{\textcolor{blue}{R#1}}{%
        \ifstrequal{#1}{3}{\textcolor{magenta}{R#1}}{%
        \ifstrequal{#1}{4}{\textcolor{teal}{R#1}}{%
                           \textcolor{cyan}{R#1}%
        }}}}%
    }%
}}

\DeclareMathOperator{\tr}{tr}  

\usepackage{xr-hyper}

\makeatletter
\newcommand*{\addFileDependency}[1]{
  \typeout{(#1)}
  \@addtofilelist{#1}
  \IfFileExists{#1}{}{\typeout{No file #1.}}
}

\makeatother

\usepackage[pagebackref,breaklinks,colorlinks]{hyperref}
\usepackage[capitalize]{cleveref}
\crefname{section}{Sec.}{Secs.}
\crefname{table}{Table}{Tables}
\crefname{figure}{Fig.}{Figs.}

\begin{document}
\title{\paperTitle}
\author{\authorBlock}
\maketitle

\begin{abstract}
Monocular 3D human pose and shape estimation is an ill-posed problem since multiple 3D solutions can explain a 2D image of a subject. Recent approaches predict a probability distribution over plausible 3D pose and shape parameters conditioned on the image. We show that these approaches exhibit a trade-off between three key properties: \textbf{(i) accuracy} - the likelihood of the ground-truth 3D solution under the predicted distribution, \textbf{(ii) sample-input consistency} - the extent to which 3D samples from the predicted distribution match the visible 2D image evidence, and \textbf{(iii) sample diversity} - the range of plausible 3D solutions modelled by the predicted distribution. Our method, HuManiFlow, predicts simultaneously accurate, consistent and diverse distributions. We use the human kinematic tree to factorise full body pose into ancestor-conditioned per-body-part pose distributions in an autoregressive manner. Per-body-part distributions are implemented using normalising flows that respect the manifold structure of SO(3), the Lie group of per-body-part poses. We show that ill-posed, but ubiquitous, 3D point estimate losses reduce sample diversity, and employ only probabilistic training losses. Code is available at: \url{https://github.com/akashsengupta1997/HuManiFlow}.

\vspace{-0.4cm}
\end{abstract}
\section{Introduction}
\label{sec:intro}

\begin{figure}
    \centering
    \includegraphics[width=0.99\linewidth]{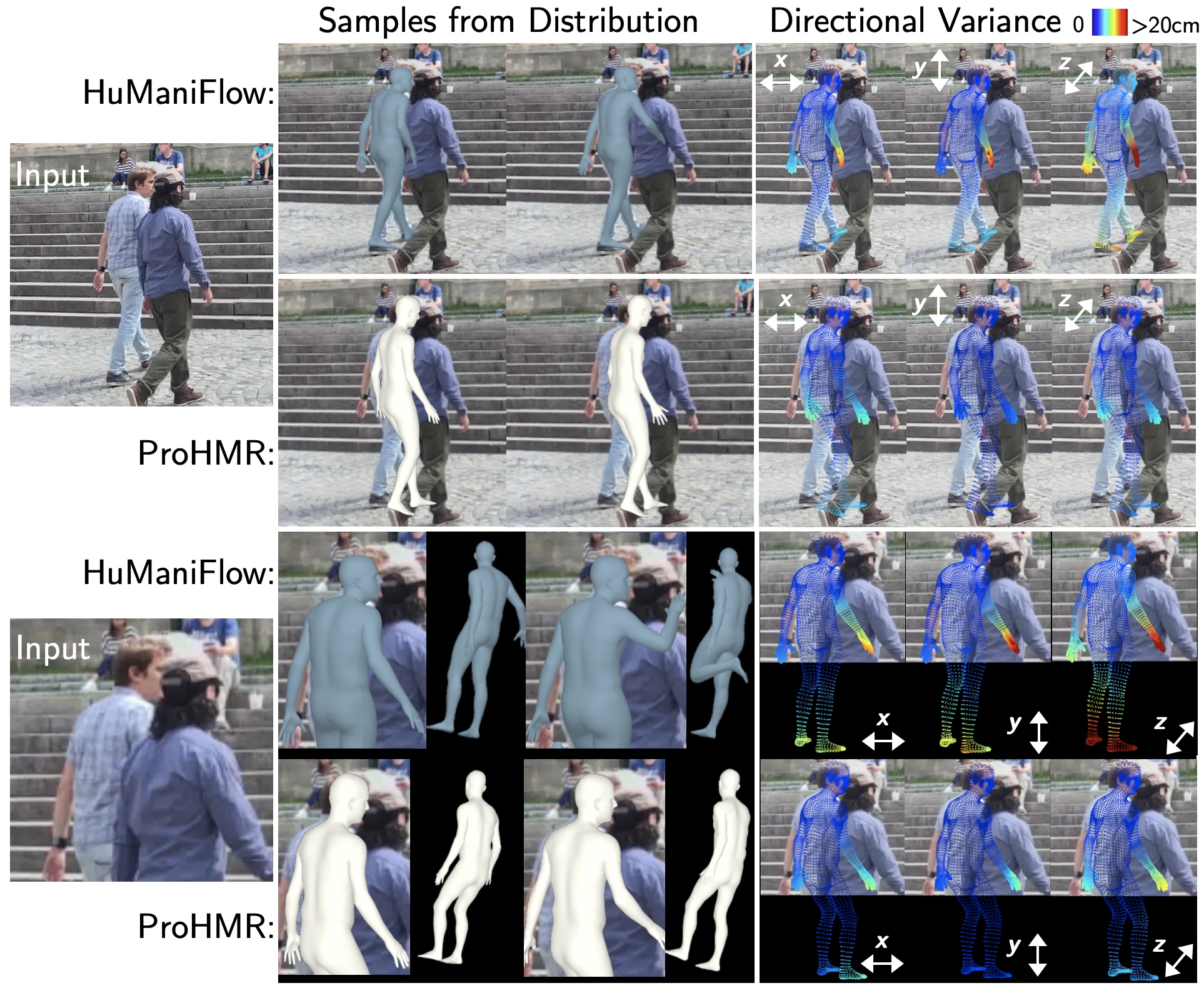}
    \caption{Comparison between pose and shape distributions from HuManiFlow and ProHMR \cite{kolotouros2021prohmr}. 3D samples from HuManiFlow are \textit{consistent} with the visible 2D evidence, while being more \textit{diverse} than samples from ProHMR. Per-vertex sample variances along the x/y/z-axes highlight interpretable uncertainty due to occlusion/truncation (all axes), and depth ambiguity (z-axis-specific).}
    \vspace{-0.4cm}
    \label{fig:intro}
\end{figure}

Estimating 3D human pose and shape from a single RGB image is an inherently ill-posed \cite{LEE1985148, Sminchisescu2003kinematicjump} computer vision task. Many 3D solutions can correspond to an input 2D observation, due to depth ambiguity, occlusion and truncation. Thus, several recent approaches \cite{sengupta2021hierprobhuman, biggs2020multibodies, kolotouros2021prohmr, sengupta2021probabilisticposeshape, aliakbarian2022flag} use deep neural networks to predict a probability \textit{distribution} over 3D pose and shape, conditioned on the 2D input. In theory, this has a few advantages over deterministic single-solution predictors \cite{hmrKanazawa17, kolotouros2019spin, Kocabas_PARE_2021, li2020hybrik, zanfir2021hund} - such as the quantification of prediction uncertainty, sampling of multiple plausible 3D solutions, and usage in downstream tasks such as multi-input fusion \cite{sengupta2021probabilisticposeshape, sengupta2021semanticlocal} or as a prior in parametric model fitting \cite{kolotouros2021prohmr}.

To fully realise the advantages of probabilistic 3D pose and shape estimation in practice, we suggest that predicted distributions should exhibit three properties: \textit{accuracy}, sample-input \textit{consistency} and sample \textit{diversity}. Accuracy denotes the likelihood of the ground-truth (GT) 3D pose and shape under the distribution. Sample-input consistency measures the extent to which 3D samples from the distribution match the 2D input. In particular, after projection to the image plane, samples should agree with any pose and shape information visible in the image. Sample diversity refers to the range of 3D poses and shapes modelled by the distribution. The GT pose and shape is but one 3D solution - the predicted distribution should model several plausible solutions when ill-posedness arises due to occlusion, truncation and depth ambiguity in the 2D input. More diverse samples enable better estimates of prediction uncertainty.

We show that recent probabilistic approaches suffer from a trade-off between accuracy, consistency and diversity. Several methods \cite{sengupta2021hierprobhuman, sengupta2021probabilisticposeshape, kolotouros2021prohmr, sengupta2021semanticlocal} predict uni-modal pose and shape distributions with limited expressiveness, and use non-probabilistic loss functions such as L1/L2 losses between GT 3D keypoints and a 3D point estimate (usually the mode of the predicted distribution). These choices favour accuracy and consistency but harm diversity, as shown in Figure \ref{fig:intro}. Approaches that generate diverse samples \cite{biggs2020multibodies}, through the use of more expressive probability distributions, often yield samples that are not consistent with the 2D input image.

We aim to balance accuracy, consistency and diversity with our approach, \textbf{HuManiFlow}, which outputs a distribution over SMPL \cite{SMPL:2015} pose and shape parameters conditioned on an input image. We use normalising flows \cite{Rezende2015normflows} to construct expressive full body pose distributions, which are factorised into per-body-part distributions autoregressively conditioned on ancestors along the human kinematic tree. We account for the manifold structure of the Lie Group of per-body-part poses (or 3D rotations) $SO(3)$ by predicting distributions on the corresponding Lie algebra $\mathfrak{so}(3)$, and ``pushing forward'' the algebra distributions onto the group via the exponential map \cite{falorsi2019reparameterizing}. We follow \cite{sengupta2021hierprobhuman, sengupta2021probabilisticposeshape} in predicting a Gaussian distribution over SMPL's shape-space PCA coefficients. Notably, our method is trained without commonly-used point estimate losses on 3D keypoints. We demonstrate that such non-probabilistic losses reduce sample diversity while providing negligible accuracy improvements when expressive distribution estimation models are used.

In summary, our main contributions are as follows:
\begin{itemize}
    \item We demonstrate that current probabilistic approaches to monocular 3D human pose and shape estimation suffer from a trade-off between distribution accuracy, sample-input consistency and sample diversity.
    
    \item We propose HuManiFlow, a normalising-flow-based method to predict distributions over SMPL pose and shape parameters that (i) considers the manifold structure of the 3D body-part rotation group $SO(3)$, (ii) exploits the human kinematic tree via autoregressive factorisation of full body pose into per-body-part rotation distributions, and (iii) is trained without any non-probabilistic point estimate losses on 3D keypoints (such as vertices or body joints).
    
    \item We show that HuManiFlow provides more accurate, input-consistent and diverse pose and shape distributions than current approaches, using the 3DPW \cite{vonMarcard2018} and SSP-3D \cite{STRAPS2020BMVC} datasets (see Figure \ref{fig:intro}). Our method interpretably and intuitively models uncertainty due to occlusion, truncation and depth ambiguities.
\end{itemize}

\section{Related Work}
\label{sec:related}

\noindent \textbf{Monocular 3D pose and shape estimation} approaches can be labelled as optimisation-based or learning-based. Optimisation-based approaches involve iteratively updating the parameters of a 3D body model \cite{SMPL:2015, SMPL-X:2019, xu2020ghum} (``model fitting'') to match 2D observations, such as 2D keypoints \cite{Bogo:ECCV:2016, Lassner:UP:2017}, silhouettes \cite{Lassner:UP:2017}, body-part masks \cite{Zanfir_2018_CVPR} or dense correspondences \cite{Guler_2019_CVPR_holopose}. These methods do not need expensive 3D-labelled training data, but require accurate 2D observations, good parameter initialisations and suitable 3D pose priors.

Learning-based approaches may be model-free or model-based. Model-free approaches directly regress a 3D human representation, such as a voxel grid \cite{varol18_bodynet}, vertex mesh \cite{kolotouros2019cmr, Moon_2020_ECCV_I2L-MeshNet, zhang2019danet, Choi_2020_ECCV_Pose2Mesh} or implicit surface \cite{saito2019pifu, saito2020pifuhd}. Model-based methods \cite{hmrKanazawa17, STRAPS2020BMVC, tan2017, zanfir2021hund, Kocabas_PARE_2021} regress body model parameters \cite{SMPL:2015, SMPL-X:2019, xu2020ghum}. Both typically use deep neural networks (DNNs).

Recently, several approaches have combined optimisation and learning. SPIN \cite{kolotouros2019spin} initialises model fitting with a DNN regression, then supervises the DNN with the optimised parameters. EFT \cite{joo2020eft} optimises the weights of a pre-trained DNN at test-time, instead of body model parameters. HybrIK \cite{li2020hybrik} uses a DNN to regress the longitudinal (twist) rotation of 3D joints, and computes the in-plane (swing) rotation analytically using inverse kinematics.

\noindent \textbf{3D pose and shape distribution estimation.} Early optimisation-based approaches to 3D pose estimation \cite{Sminchisescu2001covsampling, Sminchisescu2002hyper, Sminchisescu2003kinematicjump, Choo2001tracking} specified multi-modal posterior probabilities of 3D pose given 2D observations, and provided methods to sample multiple plausible 3D poses from the posterior. Recent learning-based approaches predict distributions over 3D keypoint locations conditioned on 2D images, using mixture density networks \cite{Bishop94mixturedensity, Oikarinen2020graphmdn, Li_2019_CVPR} or normalising flows \cite{Wehrbein2021posenormflows}. Other methods extend this to distributions over 3D pose represented by body-part rotations. 3D Multibodies \cite{biggs2020multibodies} predicts a categorical distribution over SMPL \cite{SMPL:2015} pose and shape parameters, while Sengupta \etal \cite{sengupta2021probabilisticposeshape} estimate a Gaussian distribution. HierProbHumans \cite{sengupta2021hierprobhuman} outputs a hierarchical matrix-Fisher distribution over body-part rotations, informed by the SMPL kinematic tree. ProHMR \cite{kolotouros2021prohmr} uses additive-coupling normalising flows \cite{dinh2015nice} to learn more-expressive distributions over SMPL pose parameters. These methods exhibit a trade-off between distribution accuracy, sample-input consistency and sample diversity, as we show in Section \ref{sec:experiments}.

\section{Method}
\label{sec:method}
This section provides preliminary overviews of normalising flows \cite{Rezende2015normflows}, the Lie group $SO(3)$ and SMPL \cite{SMPL:2015}, then details our pose and shape distribution prediction method.

\begin{figure*}[t]
    \centering
    \includegraphics[width=0.88\linewidth]{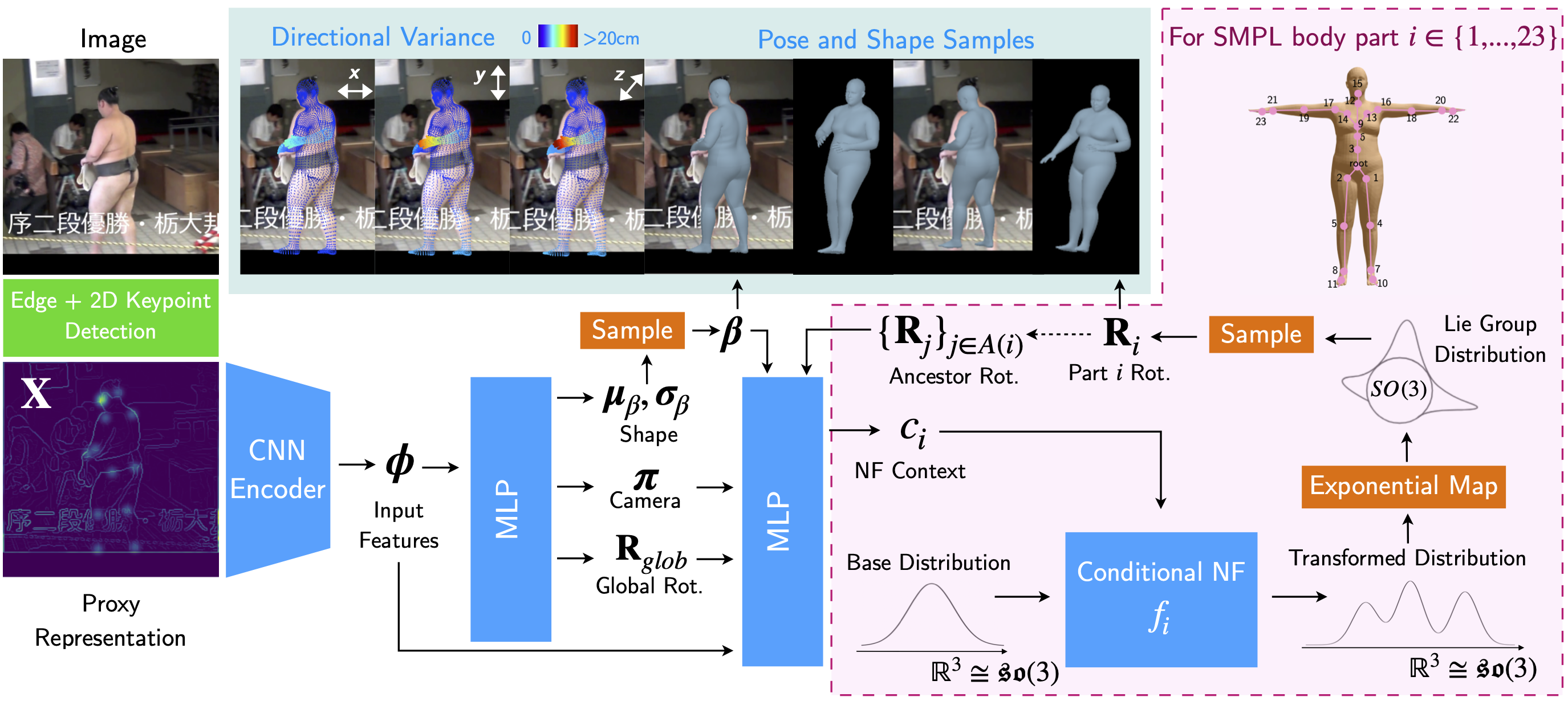}
    \caption{Overview of HuManiFlow, our method for image-conditioned pose and shape distribution prediction. We output a Gaussian distribution over SMPL \cite{SMPL:2015} shape, and learn normalising flows \cite{Rezende2015normflows, dolatabadi2020lrs} over full body pose, factorised into  ancestor-conditioned per-body-part distributions using the SMPL kinematic tree. Normalising flows are defined on $SO(3)$, the Lie group of per-body-part poses, by pushing distributions over the Lie algebra $\mathfrak{so}(3) \cong \mathbb{R}^3$ through the exponential map \cite{falorsi2019reparameterizing}. 3D body samples are obtained using ancestral sampling.}
    \label{fig:method}
    \vspace{-0.4cm}
\end{figure*}

\subsection{Normalising flows}
\label{subsec:normflows}
Normalising flows \cite{Rezende2015normflows, JMLR:papamak:v22:19-1028} are a method for constructing expressive probability distributions using the change-of-variables formula. Specifically, pushing a $D$-dimensional continuous random vector $Z \sim p_Z(\mathbf{z})$ through a diffeomorphism $f: \mathbb{R}^D \rightarrow \mathbb{R}^D$ (i.e. a bijective differentiable map with differentiable inverse $f^{-1}$) induces a random vector $Y = f(Z)$ with density function given by
\begin{equation}
\label{eqn:change_of_var}
\begin{aligned}
    p_Y(\mathbf{y}) &= p_Z(f^{-1}(\mathbf{y})) |\det J_{f^{-1}}(\mathbf{y})|\\
           &= p_Z(\mathbf{z}) |\det J_f(\mathbf{z})|^{-1}
\end{aligned}
\end{equation}
where  $\mathbf{y} \in \mathbb{R}^D$, $\mathbf{z} = f^{-1}(\mathbf{y})$ and $J_f(\mathbf{z}) \in \mathbb{R}^{D \times D}$ is the Jacobian matrix representing the differential of $f$ at $\mathbf{z}$. Intuitively, $|\det J_f(\mathbf{z})|$ gives the relative change of volume of an infinitesimal neighbourhood around $\mathbf{z}$ due to $f$.

In a normalising flow model, $f$ is the composition of multiple simple diffeomorphisms $f = f_K \circ \ldots \circ f_1$.
Each $f_k$ is implemented using a deep neural network. Typically, the base distribution $p_Z(\mathbf{z})$ is specified as $\mathcal{N}(\mathbf{0}, \mathbf{I})$. 

This formulation may be extended to model conditional probability distributions \cite{winkler2019learning} $p_{Y|C}(\mathbf{y}|\mathbf{c})$, where $\mathbf{c} \in \mathbb{R}^C$ is a context vector, using a transformation $f: \mathbb{R}^D \times \mathbb{R}^C \rightarrow \mathbb{R}^D$ such that $\mathbf{y} = f(\mathbf{z}; \mathbf{c})$. $f$ is bijective in $\mathbf{y}$ and $\mathbf{z}$.

\subsection{Lie group structure of $SO(3)$}
\label{subsec:liegroup}
The Lie group of 3D rotations may be defined as $SO(3) = \{\mathbf{R} \in \mathbb{R}^{3\times3} | \mathbf{R}^T\mathbf{R} = \mathbf{I}, \det \mathbf{R} = 1\}$. The corresponding Lie algebra $\mathfrak{so}(3)$ (i.e. tangent space at the identity $\mathbf{I}$) consists of $3 \times 3$ skew-symmetric matrices. Since $\mathfrak{so}(3)$ is a real 3D vector space, an isomorphism from $\mathbb{R}^3$ to $\mathfrak{so}(3)$ may be defined by the hat operator $\hat{.} : \mathbb{R}^3 \rightarrow  \mathfrak{so}(3)$ where
\begin{equation}
\label{eqn:hat_operator}
\hat{\mathbf{v}} =
\begin{bmatrix}
0 & -v_3 & v_2\\
v_3 & 0 & -v_1\\
-v_2 & v_1 & 0
\end{bmatrix} \in \mathfrak{so}(3)
\end{equation}
for $\mathbf{v} = [v_1, v_2, v_3]^T \in \mathbb{R}^3$.


$SO(3)$ is a matrix Lie group; thus, the exponential map $\exp: \mathfrak{so}(3) \rightarrow SO(3)$ coincides with the matrix exponential $\exp{\hat{\mathbf{v}}} = \sum_{k=0}^\infty \frac{\hat{\mathbf{v}}^k}{k!}$ for $\hat{\mathbf{v}} \in \mathfrak{so}(3)$. In practice, we use
\begin{equation}
    \exp{\hat{\mathbf{v}}} = I + (\sin{\theta}) \hat{\mathbf{u}} + (1-\cos{\theta}) \hat{\mathbf{u}}^2
\end{equation}
i.e. the Rodrigues' rotation formula with $\hat{\mathbf{v}} = \theta\hat{\mathbf{u}}$. Here, $\theta \in \mathbb{R}$ is the rotation angle and the unit vector $\mathbf{u} \in \mathbb{R}^3$ is the axis. Thus, there is a correspondence between axis-angle vectors $\mathbf{v} = \theta \mathbf{u} \in \mathbb{R}^3$ and Lie algebra elements $\hat{\mathbf{v}} = \theta\hat{\mathbf{u}} \in \mathfrak{so}(3)$.

The exponential map for $SO(3)$ is surjective, as $SO(3)$ is connected and compact, and smooth. However, it is not injective, and thus not a diffeomorphism. This is because 
\begin{equation}
\label{eqn:surjective_exponential}
\exp(\theta\hat{\mathbf{u}}) = \exp((\theta + 2 \pi k)\hat{\mathbf{u}}) = \exp(\theta_k\hat{\mathbf{u}})
\end{equation}
for any $\theta\hat{\mathbf{u}} \in \mathfrak{so}(3)$, $k \in \mathbb{Z}$ and $\theta_k = \theta + 2 \pi k$. Given a rotation $\mathbf{R} \in SO(3)$, the $\log$ operator may be used to find the corresponding axis-angle vector with the smallest angle (or minimum 2-norm). Specifically, $\theta \mathbf{u} = \log{\mathbf{R}}$ such that
\begin{equation}
    \theta = \arccos\left(\frac{\tr \mathbf{R} - 1}{2}\right), \; 
    \mathbf{u} = \frac{1}{2\sin\theta} 
      \begin{bmatrix}
      R_{32} - R_{23}\\
      R_{13} - R_{31}\\
      R_{21} - R_{12}
     \end{bmatrix} 
\end{equation}
where $\theta \in [0, \pi]$. The set of all equivalent axis-angle vectors, or $\mathfrak{so}(3)$ elements, can be obtained with Eqn. \ref{eqn:surjective_exponential} - thus providing a many-valued inverse function to $\exp$.

\begin{figure*}[t!]
    \centering
    \includegraphics[width=0.95\linewidth]{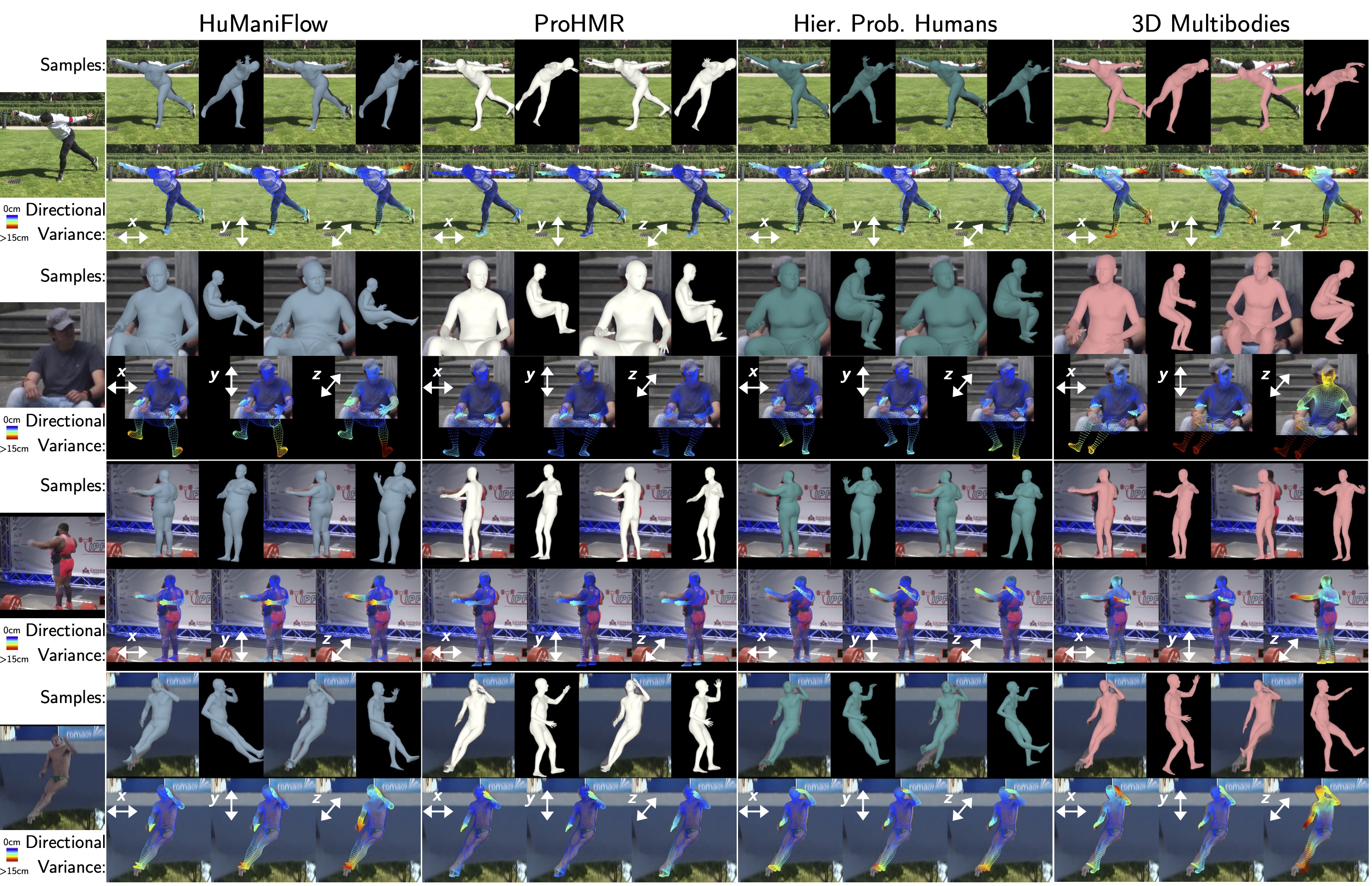}
    \caption{Comparison between HuManiFlow and recent probabilistic approaches to monocular pose and shape estimation. 3D Multibodies \cite{biggs2020multibodies} generates diverse 3D samples, but these do not consistently match the input image. Samples from ProHMR \cite{kolotouros2021prohmr} and HierProbHumans \cite{sengupta2021hierprobhuman} are input-consistent but not diverse, and tend to cluster around the predicted distribution's mode. Our method outputs consistent and diverse samples, and interpretably captures uncertainty due to depth ambiguity (z-axis variance), truncations and occlusions (all-axes variance).}
    \label{fig:comparison_fig}
    \vspace{-0.3cm}
\end{figure*}

\subsection{Constructing distributions on $SO(3)$}

We ultimately aim to learn expressive probability distributions on the Lie group $SO(3)$. We can construct a distribution on the Lie algebra $\mathfrak{so}(3)$ 
using a distribution over axis-angle vectors in $\mathbb{R}^3$, since $\mathbb{R}^3$  is isomorphic to $\mathfrak{so}(3)$ (or $\mathbb{R}^3 \cong \mathfrak{so}(3)$). The axis-angle density function, $p_{\mathbb{R}^3}(\mathbf{v})$ for $\mathbf{v} \in \mathbb{R}^3$, can be modelled by, for example, normalising flows (NFs) \cite{Rezende2015normflows} or mixture density networks (MDNs) \cite{Bishop94mixturedensity}.

$p_{\mathbb{R}^3}(\mathbf{v})$ is pushed from $\mathbb{R}^3 \cong \mathfrak{so}(3)$ onto $SO(3)$ via the exponential map, using a variant of the change-of-variables formula in Eqn. \ref{eqn:change_of_var}. $\exp$ is not a diffeomorphism over all of $\mathfrak{so}(3)$, which obstructs change-of-variables. However, it is a \textit{local} diffeomorphism (and surjective) within an open Euclidean ball $B_r(\mathbf{0})$ of radius $\pi < r < 2\pi$ about $\mathbf{0} \in \mathbb{R}^3 \cong \mathfrak{so}(3)$, as shown by \cite{falorsi2019reparameterizing}, who then use this property to derive a change-of-variables formula for the $\exp$ map:
\begin{equation}
    p_{SO(3)}(\mathbf{R}) = \sum_{k\in\{0,\pm1\}} p_{\mathbb{R}^3} (\theta_k \mathbf{u}) |\det J_\text{exp}(\theta_k \mathbf{u})|^{-1}
\label{eqn:exp_change_of_var}
\end{equation}
where $\mathbf{R} \in SO(3)$, $\theta \mathbf{u} = \mathbf{v} = \log{\mathbf{R}}$ and $\theta_k = \theta + 2 \pi k$. $p_{\mathbb{R}^3}$ must be constructed with \textit{compact support}, such that $p_{\mathbb{R}^3}(\mathbf{v}) = 0$ for $\mathbf{v} \notin B_r(\mathbf{0})$, to allow for a finite sum  over $k \in \{0, \pm1\}$ instead of $k \in \mathbb{Z}$ (as in Eqn. \ref{eqn:surjective_exponential}). $p_{SO(3)}$ is the density function on $SO(3)$ induced by pushing $p_{\mathbb{R}^3}$ through $\exp$. The determinant of $J_\text{exp}$ at $\mathbf{v} = \theta \mathbf{u}$ is given by \cite{falorsi2019reparameterizing}
\begin{equation}
    \det J_\text{exp}(\mathbf{v}) = \det J_\text{exp}(\theta \mathbf{u}) = \frac{2-2\cos\theta}{\theta^2}
\end{equation}
which, intuitively, gives the relative change of volume due to $\exp$ of an infinitesimal neighbourhood around $\mathbf{v} = \theta \mathbf{u}$. 


\subsection{SMPL model}
\label{subsec:smpl}
SMPL \cite{SMPL:2015} is a parametric human body model. Body shape is parameterised by coefficients $\boldsymbol{\beta} \in \mathbb{R}^{10}$ corresponding to a PCA shape-space basis. Pose is given by the 3D rotation of each body-part relative to its parent joint in the kinematic tree, which consists of 23 body (i.e. non-root) joints. Full body pose can be represented as a set of per-body-part relative rotations $\{\mathbf{R}_i\}_{i=1}^{23}$ where $\mathbf{R}_i \in SO(3)$. We denote the global rotation about the root joint as $\mathbf{R}_{glob}$. SMPL provides a function $\mathcal{M}(\boldsymbol{\beta}, \{\mathbf{R}_i\}_{i=1}^{23}, \mathbf{R}_\text{glob})$ that outputs a vertex mesh $\mathbf{V} \in \mathbb{R}^{6890 \times 3}$. 3D keypoints are given by $\mathbf{J}^\text{3D} = \mathcal{J}\mathbf{V}$ where $\mathcal{J}$ is a pre-trained linear regressor.

\begin{table*}[t!]
\centering
\footnotesize
\renewcommand{\tabcolsep}{1.6pt}
\begin{tabular}{l l | c c c | c c c} 
\hline
\multirow{2}{0.14\linewidth}{\textbf{Body Pose Distribution On}} & \multirow{2}{0.1\linewidth}{\textbf{Distribution Type}} & \multicolumn{3}{c|}{\textbf{3DPW}} & \multicolumn{3}{c}{\textbf{3DPW Cropped }} \\
& & \textit{Accuracy} & \textit{Consistency} & \textit{Diversity} & \textit{Accuracy} & \textit{Consistency} & \textit{Diversity} \\
& & MPJPE-PA & 2DKP Error & 3DKP Spread & MPJPE-PA & 2DKP Error & 3DKP Spread\\
& & Point / Sample Min. & Point / Samples & Vis. / Invis. & Point / Sample Min. & Point / Samples & Vis. / Invis.\\
\hline
\hline
\multirow{3}{0.17\linewidth}{\textit{Full Body (concatenated) Axis-Angles }$\boldsymbol{\Theta} \in \mathbb{R}^{69}$} & Gaussian & 55.3 / 47.8 \footnotesize$(\text{13.6\%})$ & 5.9 / 8.2 & 48.2 / 118.6 & 78.8 / 62.8 \footnotesize$(\text{20.3\%}$) & 10.9 / 15.5  & 57.4 / 140.7 \\
 & MDN \cite{Bishop94mixturedensity} & 54.6 / 46.9 \footnotesize$(\text{14.1\%})$ & 5.9 / 8.1 & 50.5 / 122.3 & 78.4 / 62.2 \footnotesize$(\text{20.7\%}$) & 10.9 / 15.5  & 57.4 / 140.7 \\
 & LRS-NF \cite{dolatabadi2020lrs} & 54.5 / 46.1 \footnotesize$(\text{15.4\%})$ & 5.8 / 8.1 & 51.3 / 124.0 & 78.6 / 61.9  \footnotesize$(\text{21.2\%})$ & 10.8 / 15.5 & 57.6 / 140.9 \\
\hline
\multirow{3}{0.17\linewidth}{\textit{Ancestor-Conditioned Axis-Angles} $\{\mathbf{v}_i\}_{i=1}^{23}$, $\mathbf{v}_i \in \mathbb{R}^3 \cong \mathfrak{so}(3)$} & Gaussian & 55.0 / 43.6 \footnotesize$(\text{20.7\%})$ & 5.4 / 6.7 & 43.2 / 105.3 & 84.8 / 60.5 \footnotesize$(\text{28.7\%})$ & 9.8 / 12.5 & 43.8 / 136.4 \\
& MDN \cite{Bishop94mixturedensity} & 54.9 / 41.7 \footnotesize$(\text{24.0\%})$ & 5.3 / 6.7 & 47.6 / 118.7 & 84.4 / 60.0 \footnotesize$(\text{28.9\%})$ & \textbf{9.7} / 12.2 & 43.7 / 138.9\\
& LRS-NF \cite{dolatabadi2020lrs} & \textbf{53.4} / 41.1 \footnotesize$(\text{23.0\%})$ & \textbf{5.1} / 6.6 & 44.7 / 110.5 & 83.6 / 59.3 \footnotesize$(\text{29.1\%}$) & 9.8 / 11.9 & 38.9 / 130.2 \\
\hline
\multirow{3}{0.17\linewidth}{\textit{Ancestor-Conditioned Matrices} $\{\mathbf{R}_i\}_{i=1}^{23}$, $\mathbf{R}_i \in SO(3)$ \textit{manifold}} & Matrix-Fisher & 54.0 / 43.4 \footnotesize$(\text{19.7\%}$) & 5.1 / 6.8 & 51.4 / \textbf{131.7} & 80.4 / 58.5 \footnotesize$(\text{27.2\%}$) & 9.9 / 11.6  & 49.6 / \textbf{142.7} \\
& MDN \cite{Bishop94mixturedensity} & 54.3 / 40.8 \footnotesize$(\text{24.9\%}$) & 5.2 / 6.7 & 47.2 / 119.2 & 80.3 / 57.8 \footnotesize$(\text{28.0\%}$) & 9.8 / 11.5 & 42.8 / 139.2 \\
& LRS-NF \cite{dolatabadi2020lrs} & \textbf{53.4} / \textbf{39.9} \footnotesize$(\textbf{25.3\%}$) & \textbf{5.1} / \textbf{6.2} & 42.8 / 116.0 & \textbf{78.2} / \textbf{54.9} \footnotesize$(\textbf{29.8\%}$) & 9.8 / \textbf{11.3} & 40.0 / 128.5  \\
\hline
\end{tabular}
\caption{Ablation study comparing pose distribution modelling choices in terms of accuracy, sample-input consistency and sample diversity metrics (see Section \ref{sec:imp_details}) on 3DPW. ``Point'' indicates point estimate metrics. MPJPE-PA and 3DKP spread are in mm, while 2DKP reprojection error is in pixels. Brackets contain $\%$ decreases from point estimate MPJPE-PA to the min. sample MPJPE-PA computed using 100 samples.}
\label{table:3dpw_ablation_distributions}
\vspace{-0.4cm}
\end{table*}

\subsection{Pose and shape distribution prediction}
\label{subsec:pose_shape_dist_pred}

Our method, HuManiFlow, predicts probability distributions over SMPL pose and shape parameters conditioned on an input image, as shown in Figure \ref{fig:method}. It also outputs deterministic estimates of weak-perspective camera parameters $\pmb{\pi} = [s, t_x, t_y]$, denoting scale and $xy$ translation, and $\mathbf{R}_\text{glob}$.

Given an input image, we first compute a proxy representation $\mathbf{X} \in \mathbb{R}^{H \times W \times C}$ consisting of an edge map and 2D keypoint heatmaps \cite{sengupta2021hierprobhuman}, stacked along the channel dimension (see Figure \ref{fig:method}). These are obtained using Canny edge detection \cite{canny1986edge} and HRNet-W48 \cite{sun2019hrnet} respectively. Proxy representations are often used to bridge the gap between synthetic training images and real test images \cite{STRAPS2020BMVC, Charles2020realtimesscreen}.

The proxy representation is passed through a CNN encoder \cite{He2015} to give input features. The camera $\pmb{\pi}$ and global rotation $\mathbf{R}_\text{glob}$ are regressed from these features with an MLP.

Next, we predict a joint distribution over SMPL pose and shape parameters, $p_\text{joint}(\{\mathbf{R}_i\}_{i=1}^{23}, \boldsymbol{\beta} | \mathbf{X})$, conditioned on the input proxy representation $\mathbf{X}$. $p_\text{joint}$ is factorised into 
\begin{equation}
    p_\text{joint}(\{\mathbf{R}_i\}_{i=1}^{23}, \boldsymbol{\beta} | \mathbf{X}) = p_\text{shape}(\boldsymbol{\beta} | \mathbf{X}) p_\text{pose}(\{\mathbf{R}_i\}_{i=1}^{23}| \boldsymbol{\beta}, \mathbf{X}).
\label{eqn:joint_dist}
\end{equation}
We condition the full body pose distribution $p_\text{pose}$ on $\boldsymbol{\beta}$ as this determines 3D body-part proportions, which affects the posed locations of mesh vertices. In practice, we also explicitly condition $p_\text{pose}$ on $\pmb{\pi}$ and $\mathbf{R}_\text{glob}$ (see Figure \ref{fig:method}). These are functions of $\mathbf{X}$, and are notationally omitted for  simplicity.

Following \cite{sengupta2021probabilisticposeshape}, we predict a Gaussian shape distribution
\begin{equation}
    p_\text{shape}(\boldsymbol{\beta} |\mathbf{X}) = \mathcal{N}(\boldsymbol{\beta}; \boldsymbol{\mu}_\beta(\mathbf{X}), 
    \text{diag}(\boldsymbol{\sigma}^2_\beta(\mathbf{X}))
\label{eqn:shape_dist}
\end{equation}
where $\boldsymbol{\mu}_\beta$ and $\boldsymbol{\sigma}_\beta^2$ are obtained with an MLP.

In SMPL, each body-part's pose is defined relative to its parent joint. The parent joint is rotated about its own parent, all the way up the kinematic tree. Thus, it is reasonable to inform the $i$-th body-part's pose $\mathbf{R}_i$ on $\{\mathbf{R}_j\}_{j\in A(i)}$, the rotations of all its kinematic ancestors $A(i)$, as noted by \cite{georgakis2020hkmr, sengupta2021hierprobhuman}. This motivates an autoregressive factorisation of $p_\text{pose}$ into ancestor-conditioned per-body-part rotation distributions:
\begin{equation}
\begin{aligned}
      p_\text{pose}(\{\mathbf{R}_i\}_{i=1}^{23}| \boldsymbol{\beta}, \mathbf{X}) 
    &= \prod_{i=1}^{23} p_{SO(3)}(\mathbf{R}_i | \{\mathbf{R}_j\}_{j\in A(i)}, \boldsymbol{\beta}, \mathbf{X})\\
    &= \prod_{i=1}^{23} p_{SO(3)}(\mathbf{R}_i | \mathbf{c}_i)
\end{aligned}
\label{eqn:pose_dist}
\end{equation}
where $\mathbf{c}_i$ is a context vector, which is computed as a function of $\{\mathbf{R}_j\}_{j\in A(i)}$, $\boldsymbol{\beta}$ and $\mathbf{X}$, as shown in Figure \ref{fig:method}. Autoregressive factorisation is similar to the hierarchical distribution proposed in \cite{sengupta2021hierprobhuman}. However, we condition part rotations directly on ancestor \textit{rotations}, instead of ancestor \textit{distribution parameters} as in \cite{sengupta2021hierprobhuman}. This enables more input-consistent distributions, since rotation samples give the \textit{exact} 3D locations of ancestor joints, while distribution parameters only say what the ancestors’ rotations are \textit{likely} to be.


We implement $p_{SO(3)}(\mathbf{R}_i | \mathbf{c}_i)$, the $i$-th body-part rotation distribution, by first defining a conditional NF over the axis-angle vector $\mathbf{v}_i\in \mathbb{R}^3 \cong \mathfrak{so}(3)$, with density function $p_{\mathbb{R}^3}(\mathbf{v}_i | \mathbf{c}_i)$. This is shown in Figure \ref{fig:method}, where the $i$-th flow diffeomorphism is denoted as $f_i: \mathbb{R}^3 \rightarrow \mathbb{R}^3$. Then, letting $\mathbf{v}_i = \theta_i \mathbf{u}_i$, Eqn. \ref{eqn:exp_change_of_var} is used to push $p_{\mathbb{R}^3}(\mathbf{v}_i | \mathbf{c}_i)$ onto $SO(3)$, finally yielding $p_{SO(3)}(\mathbf{R}_i | \mathbf{c}_i)$ via the exponential map. 

Eqn. \ref{eqn:exp_change_of_var} requires $p_{\mathbb{R}^3}(\mathbf{v}_i | \mathbf{c}_i)$ to have compact support within $B_r(\mathbf{0})$ with $\pi < r < 2\pi$. Thus, we implement a bijective radial tanh transform \cite{falorsi2019reparameterizing} $t: \mathbb{R}^3 \rightarrow B_r(\mathbf{0})$, where
\begin{equation}
\label{eqn:radial_tanh}
    t(\mathbf{x}) = r \tanh\left(\frac{\|\mathbf{x}\|}{r}\right)\frac{\mathbf{x}}{\|\mathbf{x}\|}, 
\end{equation}
 as the last layer of each $f_i$. We improve the transform proposed in \cite{falorsi2019reparameterizing}, by ensuring that $t(\mathbf{x}) \approx \mathbf{x}$ for small $\|\mathbf{x}\|$, which empirically aids training as shown in the supplement.

\begin{table*}[t]
\centering
\footnotesize
\renewcommand{\tabcolsep}{3.2pt}
\begin{tabular}{c c c | c c c | c c c} 
\hline
\multicolumn{3}{c|}{\textbf{Losses Used}} & \multicolumn{3}{c|}{\textbf{3DPW}} & \multicolumn{3}{c}{\textbf{3DPW Cropped }} \\
& & & \textit{Accuracy} & \textit{Consistency} & \textit{Diversity} & \textit{Accuracy} & \textit{Consistency} & \textit{Diversity} \\
NLL & \multirow{2}{0.08\linewidth}{2DKP Samples} & \multirow{2}{0.12\linewidth}{3DKP + 3D Vert. Point Estimate} & MPJPE-PA & 2DKP Error & 3DKP Spread & MPJPE-PA & 2DKP Error & 3DKP Spread\\
& & & Point / Sample Min. & Point / Samples & Vis. / Invis. & Point / Sample Min. & Point / Samples & Vis. / Invis.\\
\hline
\hline
\checkmark & & & 56.3 / 42.3 \footnotesize$(\text{24.9\%}$) & 5.6 / 8.1 & 55.7 / \textbf{124.0}  & 90.2 / 64.9 \footnotesize$(\text{28.0\%}$) & 11.0 / 15.1 & 66.3 / \textbf{141.0} \\
\checkmark & \checkmark & & 53.4 / \textbf{39.9} \footnotesize$(\textbf{25.3\%}$) & \textbf{5.1} / \textbf{6.2} & 42.8 / 116.0 & \textbf{78.2} / \textbf{54.9} \footnotesize$(\textbf{29.8\%}$) & \textbf{9.8}/ \textbf{11.3} & 40.0 / 128.5  \\
\checkmark & \checkmark & \checkmark & \textbf{53.3} / \textbf{39.9} \footnotesize$(\text{25.1\%}$) & \textbf{5.1} / 6.3 & 39.3 / 109.6 & 83.4 / 61.1 \footnotesize$(\text{26.7\%}$ & \textbf{9.8} / 11.4  & 39.7 / 115.1\\
\hline
\end{tabular}
\caption{Ablation study comparing probabilistic losses - i.e. negative log-likelihood and visibility-masked 2DKP samples loss - and non-probabilistic point estimate losses on 3D keypoints and vertices. Distribution accuracy, sample-input consistency and sample diversity metrics are detailed in Section \ref{sec:imp_details}. ``Point'' indicates point estimate metrics. MPJPE-PA and 3DKP spread are in mm, while 2DKP reprojection error is in pixels. Brackets contain $\%$ decreases from point estimate MPJPE-PA to the min. sample MPJPE-PA computed using 100 samples.}
\label{table:3dpw_ablation_losses}
\vspace{-0.3cm}
\end{table*}

\subsection{Pose and shape sampling and point estimation}
SMPL pose and shape samples can be obtained from $p_\text{joint}(\{\mathbf{R}_i\}_{i=1}^{23}, \boldsymbol{\beta} | \mathbf{X})$ via ancestral sampling, as shown in Figure \ref{fig:method}. Specifically, we first sample $\boldsymbol{\beta} \sim p_\text{shape}(\boldsymbol{\beta} |\mathbf{X})$, which will be used to condition $p_\text{pose}$. Then, each body-part's rotation $\mathbf{R}_i \sim p_{SO(3)}(\mathbf{R}_i | \mathbf{c}_i)$ is sampled following the corresponding ``limb'' of the kinematic tree, by first sampling all ancestor rotations $\{\mathbf{R}_j\}_{j\in A(i)}$, then obtaining the context vector $\mathbf{c}_i$ from $\{\mathbf{R}_j\}_{j\in A(i)}$, $\boldsymbol{\beta}$ and $\mathbf{X}$ using an MLP. Pose and shape samples are converted into 3D vertex mesh samples with the SMPL function $\mathcal{M}$. The variance of each vertex along the x, y and z directions highlights the uncertainty captured by the predicted distribution $p_\text{joint}$, arising due to depth (i.e. z-axis) ambiguity, occlusion and truncation.

To obtain a point estimate of SMPL pose and shape given an input image, we would want to compute the mode $(\{\mathbf{R}_i^*\}_{i=1}^{23}, \boldsymbol{\beta}^*) = \text{argmax} \: p_\text{joint}(\{\mathbf{R}_i\}_{i=1}^{23}, \boldsymbol{\beta} | X)$. This is challenging since $p_\text{joint}$ is the product of multiple complex NF distributions. The mode of each per-body-part NF distribution is itself non-trivial. As an approximation, we use
\begin{equation}
    \boldsymbol{\beta}^* = \boldsymbol{\mu}_\beta, \; \mathbf{R}_i^* = \exp(f_i(\mathbf{0}; \mathbf{c}_i))
\label{eqn:point_estimate}
\end{equation}
where the $i$-th body-part's pose estimate $\mathbf{R}^*_i$ is acquired by passing the base distribution mode ($\mathbf{0}$) through the $i$-th flow transform and $\exp$ map. While the resulting $(\{\mathbf{R}_i^*\}_{i=1}^{23},\boldsymbol{\beta}^*)$  is not, in general, the mode of $p_\text{joint}$,  we show that it typically has high likelihood under $p_\text{joint}$ in the supplement.

\subsection{Loss functions}
\label{subsec:loss_functions}
We train our model using a dataset of synthetic inputs paired with ground-truth pose, shape and global rotation labels $\{\mathbf{X}^n, \{\bar{\mathbf{R}}^n\}_{i=1}^{23}, \bar{\boldsymbol{\beta}}^n, \bar{\mathbf{R}}_\text{glob}^n\}_{n=1}^N$, as discussed in Section \ref{sec:imp_details}. We apply a negative log-likelihood (NLL) loss

\begin{equation}
    \mathcal{L}_\text{NLL} = - \sum_{n=1}^N \ln p_\text{joint}(\{\bar{\mathbf{R}}_i^n\}_{i=1}^{23}, \bar{\boldsymbol{\beta}}^n | \mathbf{X}^n)
\end{equation}
over pose and shape parameters. $\mathbf{R}_\text{glob}$ is supervised using 
\begin{equation}
    \mathcal{L}_\text{glob} = \sum_{n=1}^N \| \mathbf{R}_\text{glob}(\mathbf{X}_n) - \bar{\mathbf{R}}_\text{glob}^n \|_{F}^2.
\end{equation}

Following \cite{sengupta2021hierprobhuman, kolotouros2021prohmr}, we apply a loss between 2D keypoint samples and \textit{visible} GT 2D keypoints, $\mathcal{L}_\text{2D}$, to encourage sample-input consistency. 2D keypoint samples are obtained by sampling pose and shape from $p_\text{joint}$, computing the corresponding 3D keypoint samples with SMPL, and projecting these onto the image plane using the predicted camera $\pmb{\pi}$.

The overall training loss is given by $\mathcal{L} = \lambda_\text{NLL} \mathcal{L}_\text{NLL} + \lambda_\text{glob}\mathcal{L}_\text{glob} + \lambda_\text{2D}\mathcal{L}_\text{2D}$ where the $\lambda$s are weights. We do not use ubiquitous, but non-probabilistic, point estimate losses on 3D keypoints, as justified by Section \ref{subsec:ablation} and Table \ref{table:3dpw_ablation_losses}.
\section{Implementation Details}
\label{sec:imp_details}

\nbf{Model architecture} We use a ResNet-18 \cite{He2015} CNN encoder. Per-body-part axis-angle probability densities $p_{\mathbb{R}^3}(\mathbf{v}_i | \mathbf{c}_i)$ are implemented with Linear Rational Spline normalising flows (LRS-NFs) \cite{dolatabadi2020lrs}. Further architecture and hyperparameter details are provided in the supplementary material.

\nbf{Synthetic training data} We adopt the same training data generation pipeline as \cite{sengupta2021hierprobhuman}, which renders synthetic proxy representation inputs $\{\mathbf{X}^n\}_{n=1}^N$ from ground-truth (GT) poses, shapes and global rotations $\{\{\bar{\mathbf{R}}_i^n\}_{i=1}^{23}, \bar{\boldsymbol{\beta}}^n, \bar{\mathbf{R}}_\text{glob}^n\}_{n=1}^N$. GT poses and global rotations are sampled from the training sets of UP-3D \cite{Lassner:UP:2017}, 3DPW \cite{vonMarcard2018} and Human3.6M \cite{h36m_pami}. GT shapes are randomly sampled from a prior Gaussian distribution. Truncation, occlusion and noise augmentations bridge the synthetic-to-real gap.

\nbf{Training details} We use Adam \cite{kingma2014adam}, with a learning rate of 1e-4 and batch size of 72, and train for 200 epochs.

\nbf{Evaluation datasets and metrics} We use the 3DPW \cite{vonMarcard2018} and SSP-3D \cite{STRAPS2020BMVC} datasets to evaluate the \textit{accuracy}, sample-input \textit{consistency} and sample \textit{diversity} of predicted pose and shape distributions. Moreover, we generate cropped versions of 3DPW and SSP-3D, resulting in more ambiguous test data to evaluate sample diversity. Cropped dataset generation details are provided in the supplementary material.

\textit{Distribution accuracy} refers to the likelihood of the GT pose and shape under the predicted distribution. We measure accuracy on 3DPW using MPJPE and MPJPE-PA computed with the \textit{minimum} error sample out of $N$ samples from the predicted distribution, where $N$ is increased from 1 to 100. If the GT pose and shape have high likelihood under the predicted distribution, we expect the minimum MPJPE over $N$ samples to improve significantly with increasing $N$. For $N=1$, the sample is obtained as a point estimate from the predicted pose and shape distribution, using Eqn. \ref{eqn:point_estimate}. Similarly, shape-specific distribution accuracy is measured on SSP-3D, using PVE-T-SC (per-T-pose-vertex-error after scale correction) computed with the minimum error sample.

\textit{Sample-input consistency} denotes the extent to which predicted 3D samples match the \textit{visible} pose and shape evidence in the input 2D image. We measure consistency on 3DPW and SSP-3D using the average reprojection error between predicted 2D keypoint (2DKP) samples projected onto the image plane and visible GT 2DKPs, averaged over 100 samples for each test image. The 17 COCO keypoint convention is used \cite{Lin2014MicrosoftCC}. We also compute the reprojection error between 2DKP point estimates (Eqn. \ref{eqn:point_estimate}) and GT 2DKPs. An input-consistent distribution should have low average sample reprojection error, close to that of the point estimate.

\textit{Sample diversity} refers to the range of 3D reconstructions modelled by the predicted distribution. We measure diversity by drawing 100 predicted samples, and computing the average 3D Euclidean distance from the mean for each 3D keypoint (3DKP), split into \textit{visible} and \textit{invisible} keypoints. A diverse distribution should exhibit significant spread in sample 3DKP locations - typically along the z-axis for visible 3DKPs (depth ambiguity) and all axes for invisible 3DKPs. This is a simplistic metric; defining a good diversity metric for high-dimensional, complex distributions is non-trivial, and a potential area for future research.

\begin{table}[t]
\centering
\small
\renewcommand{\tabcolsep}{3.3pt}
\begin{tabular}{l | c c | c c} 
\hline
\textbf{Method} & \multicolumn{4}{c}{\textbf{3DPW} - \textit{Accuracy}}\\
& \multicolumn{2}{c|}{MPJPE (mm)} & \multicolumn{2}{c}{MPJPE-PA (mm)}\\ 
& Point & Sample Min. & Point & Sample Min. \\
\hline
\hline
HMR \cite{hmrKanazawa17} & 130.0 & - & 76.7 & -\\
SPIN \cite{kolotouros2019spin} & 96.9 & - & 59.0  & -\\
I2L-MeshNet \cite{Moon_2020_ECCV_I2L-MeshNet} & 93.2  & - & 57.7 & -\\
DaNet \cite{zhang2019danet} & 85.5 & - & 54.8 & - \\
HUND \cite{zanfir2021hund} & 81.4 & - & 57.5 & -\\
PARE \cite{Kocabas_PARE_2021} & 74.5 & - & 46.5 & -\\
HybrIK \cite{li2020hybrik} & \textbf{74.1} & - & \textbf{45.0} & -\\
\hline
3D Multibodies \cite{biggs2020multibodies} & 93.8 & 74.6 \footnotesize$(\text{20.5\%})$ & 59.9 & 48.3 \footnotesize$(\text{19.4\%})$\\ 
Sengupta \etal \cite{sengupta2021probabilisticposeshape} & 97.1 & 84.4 \footnotesize$(
\text{13.1\%})$ & 61.1 & 52.1 \footnotesize$(\text{14.7\%})$\\
ProHMR \cite{kolotouros2021prohmr} & 97.0 & 81.5 \footnotesize$(\text{16.0\%})$ & 59.8 & 48.2 \footnotesize$(\text{19.4\%})$\\
HierProbHuman \cite{sengupta2021hierprobhuman} & 84.9 & 70.9 \footnotesize$(\text{16.5\%})$ & 53.6 & 43.8 \footnotesize$(\text{18.3\%})$\\
HuManiFlow & \textbf{83.9} & \textbf{65.1} \footnotesize$(\textbf{22.4\%})$ & \textbf{53.4} & \textbf{39.9} \footnotesize$(\textbf{25.3\%})$\\
\hline
\end{tabular}
\vspace{-0.2cm}
\caption{Comparison of recent deterministic (top) and probabilistic (bottom) methods in terms of 3D point estimate and distribution accuracy on 3DPW \cite{vonMarcard2018}.}
\label{table:3dpw_sota_comparison_accuracy}
\vspace{-0.4cm}
\end{table}

\begin{figure}[t]
    \centering
    \includegraphics[width=\linewidth]{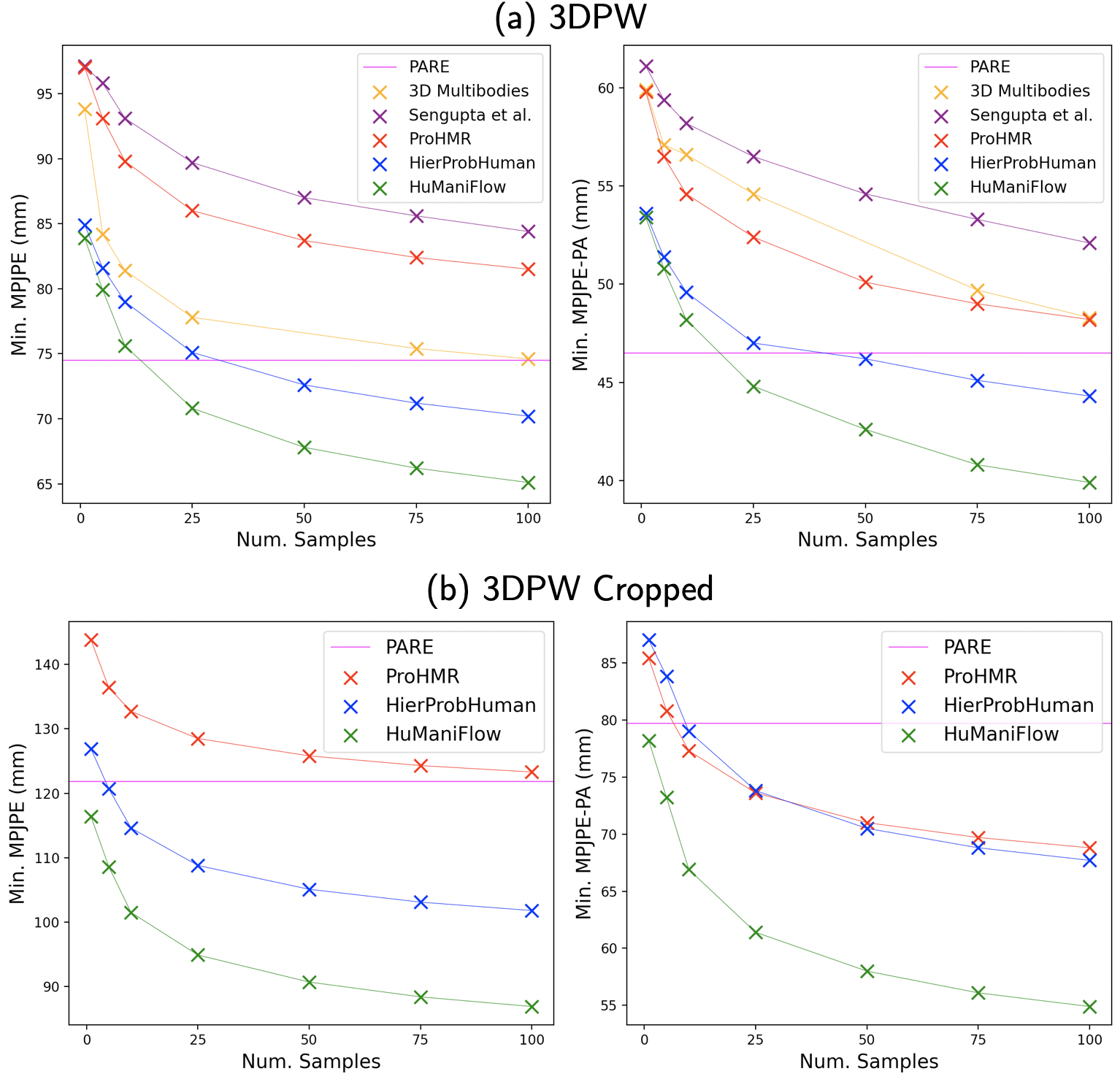}
    \caption{Rate of improvement in min. sample MPJPE(-PA) with increasing number of samples. A faster improvement rate indicates that the GT solution has higher likelihood under the predicted distribution - i.e. better distribution accuracy (see Section \ref{sec:imp_details}).}
    \label{fig:min_sample_mpjpe_3dpw}
    \vspace{-0.4cm}
\end{figure}

\section{Experimental Results}
\label{sec:experiments}

This section first presents ablation studies on distribution predictors and losses, and then compares our method's accuracy, consistency and diversity with the state-of-the-art.

\subsection{Ablation studies}
\label{subsec:ablation}

\nbf{Pose distribution modelling choices} Table \ref{table:3dpw_ablation_distributions} compares several different pose distribution prediction models, in terms of accuracy, consistency and diversity on 3DPW. Rows 1-3 report metrics from a naive approach, where SMPL pose parameters are simply treated as a vector $\boldsymbol{\Theta} \in \mathbb{R}^{69}$ formed by concatenating per-body-part axis-angle vectors $\mathbf{v}_i \in \mathbb{R}^3$. The distribution over $\boldsymbol{\Theta}$ is modelled as a multivariate Gaussian, MDN \cite{Bishop94mixturedensity} or LRS-NF \cite{dolatabadi2020lrs}. Notably, all three models perform similarly, despite the greater theoretical expressiveness of MDNs and LRS-NFs. This suggests that it is challenging to directly predict complex probability distributions over high-dimensional full body pose. The shortcomings of high-dimensional MDNs in particular are well known \cite{MICB19, rupprecht2017learning}.

Rows 4-6 in Table \ref{table:3dpw_ablation_distributions} investigate an autoregressive approach, where the full body pose distribution is factorised into per-body-part distributions on axis-angle vectors $\mathbf{v}_i$, similar to Eqn. \ref{eqn:pose_dist}. The axis-angle distributions are modelled as a 3D Gaussian, MDN or LRS-NF. However, they are not pushed onto $SO(3)$ using the change-of-variables formula for the $\exp$ map (Eqn. \ref{eqn:exp_change_of_var}). Thus, the manifold structure of $SO(3)$ and local ``change of volume'' due to $\exp$ are disregarded. These models improve consistency and accuracy over the naive approach (Rows 1-3), shown by better average sample 2DKP error and minimum sample MPJPE-PA. Despite the latter, they yield worse point estimate MPJPE-PA on the 3DPW Cropped split. This suggests that the point estimate may not be close to the GT, even for an accurate distribution, when faced with highly ambiguous inputs. The naive approach has greater sample diversity - but this is meaningless if the samples are not input-consistent.

Rows 7-9 in Table \ref{table:3dpw_ablation_distributions} investigate autoregressive per-body-part distributions over $SO(3)$ (Eqn. \ref{eqn:pose_dist}). These are defined by pushing either MDNs or LRS-NFs over $\mathbb{R}^3 \cong \mathfrak{so}(3)$ through $\exp$ onto $SO(3)$, or as a matrix-Fisher distribution \cite{down1972orientationstatistics, mardia_jupp_2000, mohlin2020matrixfisher}. Autoregressive LRS-NFs on $SO(3)$ yield the best accuracy and consistency metrics, with only slightly reduced diversity; thus, we use this model in HuManiFlow.

\nbf{Distribution prediction losses} Table \ref{table:3dpw_ablation_losses} explores probabilistic losses - i.e. NLL and a loss on 2DKP \textit{samples} (see Section \ref{subsec:loss_functions}) - and non-probabilistic losses (e.g. MSE) applied between GT 3DKPs and point estimates (Eqn. \ref{eqn:point_estimate}). Despite their ubiquity, we find that 3D point estimate losses do not improve distribution accuracy (i.e. min. Sample MPJPE-PA), or even point estimate MPJPE-PA, which is actually worsened on 3DPW Cropped. This is likely due to the ill-posedness of such losses for ambiguous inputs, where the GT 3DKPs represent but one of many plausible solutions. Point estimate losses also reduce sample diversity. Using NLL and a visibility-masked loss on 2DKP samples results in the best overall performance. Omitting the 2DKP samples loss degrades sample-input consistency, as expected.

\subsection{Comparison with the state-of-the-art}

\nbf{Distribution accuracy} Tables \ref{table:3dpw_sota_comparison_accuracy} and \ref{table:ssp3d_sota_comparison_accuracy} evaluate the accuracy of current pose and shape estimators. HuManiFlow predicts more accurate distributions (i.e. lower min. sample metrics) than other probabilistic approaches. This is corroborated by Figure \ref{fig:min_sample_mpjpe_3dpw}, where HuManiFlow has the fastest decrease in min. sample MPJPE as the number of samples is increased. On 3DPW Cropped and SSP-3D, HuManiFlow's point estimates have lower error than the deterministic SOTA approaches.

\nbf{Sample-input consistency and sample diversity} Table \ref{table:3dpw_sota_comparison_diversity_consistency} compares the consistency and diversity of current probabilistic methods. Samples from \cite{sengupta2021hierprobhuman, kolotouros2021prohmr} are input-consistent but not diverse, likely due to the use of unimodal pose distributions and 3D point estimate losses. \cite{biggs2020multibodies} generates diverse samples but does not always match the input. Our method is generally the most input-consistent, with reasonably diverse samples. A qualitative demonstration is given in Figure \ref{fig:comparison_fig}.

\begin{table}[t]
\centering
\small
\renewcommand{\tabcolsep}{3.3pt}
\begin{tabular}{l | c c l} 
\hline
\textbf{Method} & \multicolumn{3}{c}{\textbf{SSP-3D} - \textit{Accuracy (Shape)}}\\
& \multicolumn{3}{c}{PVE-T-SC (mm)} \\ 
& Point & Sample Min. & Multi-Input \\
\hline
\hline
SPIN \cite{kolotouros2019spin} & 22.2 & - & 21.9 \scriptsize{(Mean)} \\
PARE \cite{Kocabas_PARE_2021} & 21.7 & - & 21.6 \scriptsize{(Mean)} \\
HybrIK \cite{li2020hybrik} & 22.9 & - & 22.8 \scriptsize{(Mean)} \\
STRAPS \cite{STRAPS2020BMVC} & 15.9 & - & 14.4 \scriptsize{(Mean)} \\
\hline
3D Multibodies \cite{biggs2020multibodies} & 22.3 & 19.2 \footnotesize$(\text{13.9\%})$ & 22.1 \scriptsize{(Mean)}\\ 
Sengupta \etal \cite{sengupta2021probabilisticposeshape} & 15.2 & 10.4 \footnotesize$(\text{31.6\%})$ & 13.3 \scriptsize{(Prob. Comb. \cite{sengupta2021probabilisticposeshape})}\\
ProHMR \cite{kolotouros2021prohmr}& 22.2 & - & 21.9 \scriptsize{(Mean)}\\
HierProbHuman \cite{sengupta2021hierprobhuman} & 13.6 & 8.7 \footnotesize$(\text{36.0\%})$ & 12.0 \scriptsize{(Prob. Comb.)} \\
HuManiFlow & \textbf{13.5} & \textbf{8.3} \footnotesize$(\textbf{39.0\%})$ & \textbf{11.9} \scriptsize{(Prob. Comb.)} \\
\hline
\end{tabular}
\caption{Comparison of recent deterministic (top) and probabilistic (bottom) methods in terms of shape accuracy on SSP-3D \cite{STRAPS2020BMVC}.}
\label{table:ssp3d_sota_comparison_accuracy}
\vspace{-0.4cm}
\end{table}

\nbf{Model fitting with an image-conditioned prior} Distributions with greater accuracy, consistency and diversity should be better for downstream tasks. An example task is model fitting \cite{Bogo:ECCV:2016} i.e. optimising SMPL point estimates to better fit observed 2DKPs. ProHMR \cite{kolotouros2021prohmr} use their predicted distribution as an \textit{image-conditioned prior} during fitting, outperforming generic pose priors \cite{Bogo:ECCV:2016, SMPL-X:2019}. Table \ref{tab:refinement_optimisation} shows that HuManiFlow surpasses \cite{kolotouros2021prohmr} in this task. Note the difference between prior-less optimisation of HuManiFlow's point estimates (row 5) versus using the predicted distribution as a prior (row 6).

\begin{table}[t]
\centering
\small
\renewcommand{\tabcolsep}{3pt}
\begin{tabular}{l l| c c } 
\hline
\textbf{Dataset} & \textbf{Method} & \textit{Consistency} & \textit{Diversity}\\
& & 2DKP Error & 3DKP Spread\\
&  & Point / Samples & Vis. / Invis.\\
\hline
\hline
\multirow{5}{0.125\linewidth}{\textbf{3DPW}} & 3D Multibodies \cite{biggs2020multibodies} & 5.2 / 7.8 & 80.1 / \textbf{126.9} \\ 
& Sengupta \etal \cite{sengupta2021probabilisticposeshape} & 5.6 / 8.1 & 48.3 / 98.8 \\
& ProHMR \cite{kolotouros2021prohmr} & 6.8 / 7.5 & 35.1 / 60.8 \\
& HierProbHuman \cite{sengupta2021hierprobhuman} & \textbf{5.1} / 7.2 & 47.6 / 101.4 \\
& HuManiFlow & \textbf{5.1} / \textbf{6.2} & 42.8 / 116.0 \\
\hline
\multirow{4}{0.125\linewidth}{\textbf{3DPW Cropped}} & Sengupta \etal \cite{sengupta2021probabilisticposeshape} & 11.5 / 17.1 & 47.8 / 96.5 \\
& ProHMR \cite{kolotouros2021prohmr} & 11.9 / 13.4 & 32.1 / 57.1 \\
& HierProbHuman \cite{sengupta2021hierprobhuman} & \textbf{9.7} / 12.8 & 38.5 / 100.2 \\
& HuManiFlow & 9.8 / \textbf{11.3} & 40.0 / \textbf{128.5} \\
\hline
\multirow{5}{0.125\linewidth}{\textbf{SSP-3D}} & 3D Multibodies \cite{biggs2020multibodies} & 5.3 / 7.8 & 80.7 / - \\ 
& Sengupta \etal \cite{sengupta2021probabilisticposeshape} & 6.2 / 8.0  & 50.1 / -  \\
& ProHMR \cite{kolotouros2021prohmr} & 6.9 / 7.6 &  36.6 / -  \\
& HierProbHuman \cite{sengupta2021hierprobhuman} & \textbf{4.8} / 6.9 & 48.5 / -  \\
& HuManiFlow & \textbf{4.8} / \textbf{6.0} & 47.3 / - \\
\hline
\multirow{4}{0.125\linewidth}{\textbf{SSP-3D Cropped}} & Sengupta \etal \cite{sengupta2021probabilisticposeshape} & 13.3 / 18.9  & 60.2 / \textbf{139.6}  \\
& ProHMR \cite{kolotouros2021prohmr} &  13.8 / 15.2  &  41.9 / 60.8  \\
& HierProbHuman \cite{sengupta2021hierprobhuman} & \textbf{10.6} / 14.1  & 58.7 / 105.5\\
& HuManiFlow & \textbf{10.6} / \textbf{13.0} & 45.3 / 134.0 \\
\hline
\end{tabular}
\caption{Comparison of recent probabilistic methods in terms of sample-input consistency and sample diversity on 3DPW \cite{vonMarcard2018} and SSP-3D \cite{STRAPS2020BMVC}. 3DKP spread is in mm and 2DKP error is in pixels.}
\label{table:3dpw_sota_comparison_diversity_consistency}
\vspace{-0.2cm}
\end{table}

\begin{table}[t]
\small
    \centering
    \begin{tabular}{l | l | c}
        \hline
        \textbf{Method} & \textbf{Prior} & \textbf{3DPW}\\
        & & MPJPE-PA (mm)\\
        \hline
        \hline
        SPIN \cite{kolotouros2019spin} + Fit & GMM \cite{Bogo:ECCV:2016}  & 66.5\\
        SPIN + Fit & VPoser \cite{SMPL-X:2019} & 70.9\\
        SPIN + EFT \cite{joo2020eft} & - & 56.6\\
        \hline
        ProHMR \cite{kolotouros2021prohmr} + Fit & \footnotesize Image-conditioned   & 55.1\\
        HuManiFlow + Fit & None &  53.4 \\
        HuManiFlow + Fit & \footnotesize Image-conditioned & \textbf{51.2}\\
        \hline
    \end{tabular}
    \caption{Evaluation of model fitting methods with different SMPL parameter priors, including image-conditioned priors from distribution prediction methods (\cite{kolotouros2021prohmr} and HuManiFlow). Fitting  does not necessarily improve 3D point estimate accuracy, despite better model-image alignment, unless image-conditioned priors are used.}
    \label{tab:refinement_optimisation}
    \vspace{-0.4cm}
\end{table}

\section{Conclusion}
\label{sec:conclusion}
This work proposes a probabilistic approach to the ill-posed problem of monocular 3D human pose and shape estimation. We show that current methods suffer from a trade-off between distribution accuracy, sample-input consistency and sample diversity, which affects their utility in downstream tasks. Our method, HuManiFlow, uses a normalising-flow-based pose distribution which (i) accounts for the manifold structure of $SO(3)$, (ii) has an autoregressive factorisation informed by the human kinematic tree, and (iii) is trained without any ill-posed 3D point estimate losses. HuManiFlow yields more accurate, consistent and diverse distributions.

\clearpage
{\small
\bibliographystyle{ieee_fullname}
\bibliography{11_references}

\begin{thebibliography}{10}\itemsep=-1pt

\bibitem{aliakbarian2022flag}
Sadegh Aliakbarian, Pashmina Cameron, Federica Bogo, Andrew Fitzgibbon, and
  Thomas~J. Cashman.
\newblock Flag: Flow-based 3d avatar generation from sparse observations.
\newblock In {\em Proceedings of the IEEE Conference on Computer Vision and
  Pattern Recognition (CVPR)}, 2022.

\bibitem{bhatnagar2019mgn}
Bharat~Lal Bhatnagar, Garvita Tiwari, Christian Theobalt, and Gerard Pons-Moll.
\newblock Multi-garment net: Learning to dress {3D} people from images.
\newblock In {\em Proceedings of the {IEEE} International Conference on
  Computer Vision ({ICCV})}, Oct 2019.

\bibitem{biggs2020multibodies}
Benjamin Biggs, S{\'{e}}bastien Erhardt, Hanbyul Joo, Benjamin Graham, Andrea
  Vedaldi, and David Novotny.
\newblock {3D} multibodies: Fitting sets of plausible {3D} models to ambiguous
  image data.
\newblock In {\em NeurIPS}, 2020.

\bibitem{Bishop94mixturedensity}
Christopher~M. Bishop.
\newblock Mixture density networks.
\newblock Technical report, 1994.

\bibitem{Bogo:ECCV:2016}
Federica Bogo, Angjoo Kanazawa, Christoph Lassner, Peter Gehler, Javier Romero,
  and Michael~J. Black.
\newblock Keep it {SMPL}: Automatic estimation of {3D} human pose and shape
  from a single image.
\newblock In {\em Proceedings of the European Conference on Computer Vision
  (ECCV)}, Oct. 2016.

\bibitem{canny1986edge}
John~F. Canny.
\newblock A computational approach to edge detection.
\newblock {\em IEEE Transactions on Pattern Analysis and Machine Intelligence
  (PAMI)}, 8(6):679--698, 1986.

\bibitem{Charles2020realtimesscreen}
J. Charles, S. Bucciarelli, and R. Cipolla.
\newblock Real-time screen reading: reducing domain shift for one-shot
  learning.
\newblock In {\em Proceedings of the British Machine Vision Conference (BMVC)},
  2020.

\bibitem{Choi_2020_ECCV_Pose2Mesh}
Hongsuk Choi, Gyeongsik Moon, and Kyoung~Mu Lee.
\newblock Pose2mesh: Graph convolutional network for {3D} human pose and mesh
  recovery from a {2D} human pose.
\newblock In {\em Proceedings of the European Conference on Computer Vision
  (ECCV)}, 2020.

\bibitem{Choo2001tracking}
Kiam Choo and D.J. Fleet.
\newblock People tracking using hybrid monte carlo filtering.
\newblock In {\em Proceedings of the IEEE International Conference on Computer
  Vision (ICCV)}, volume~2, pages 321--328 vol.2, 2001.

\bibitem{clevert2016elu}
Djork-Arné Clevert, Thomas Unterthiner, and Sepp Hochreiter.
\newblock Fast and accurate deep network learning by exponential linear units
  (elus), 2016.

\bibitem{dinh2015nice}
Laurent Dinh, David Krueger, and Yoshua Bengio.
\newblock {NICE}: Non-linear independent components estimation.
\newblock In {\em ICLR Workshop Track}, 2015.

\bibitem{dinh2017realnvp}
Laurent Dinh, Jascha Sohl-Dickstein, and Samy Bengio.
\newblock Density estimation using real {NVP}.
\newblock In {\em Proceedings of the International Conference on Learning
  Representations (ICLR)}, 2017.

\bibitem{dolatabadi2020lrs}
Hadi~Mohaghegh Dolatabadi, Sarah Erfani, and Christopher Leckie.
\newblock Invertible generative modeling using linear rational splines.
\newblock In {\em The 23rd International Conference on Artificial Intelligence
  and Statistics (AISTATS)}, pages 4236--4246, 2020.

\bibitem{down1972orientationstatistics}
Thomas~D. Downs.
\newblock {Orientation statistics}.
\newblock {\em Biometrika}, 59(3):665--676, 12 1972.

\bibitem{falorsi2019reparameterizing}
L. Falorsi, P. de Haan, T.R. Davidson, and P. Forr{\'e}.
\newblock Reparameterizing distributions on lie groups.
\newblock {\em 22nd International Conference on Artificial Intelligence and
  Statistics (AISTATS-19)}, 2019.

\bibitem{georgakis2020hkmr}
Georgios Georgakis, Ren Li, Srikrishna Karanam, Terrence Chen, Jana Kosecka,
  and Ziyan Wu.
\newblock Hierarchical kinematic human mesh recovery.
\newblock In {\em Proceedings of the European Conference on Computer Vision
  (ECCV)}, 2020.

\bibitem{Guler_2019_CVPR_holopose}
Riza~Alp Guler and Iasonas Kokkinos.
\newblock Holopose: Holistic {3D} human reconstruction in-the-wild.
\newblock In {\em Proceedings of the IEEE Conference on Computer Vision and
  Pattern Recognition (CVPR)}, June 2019.

\bibitem{Guler2018DensePose}
Riza~Alp G\"uler, Natalia Neverova, and Iasonas Kokkinos.
\newblock Densepose: Dense human pose estimation in the wild.
\newblock In {\em Proceedings of IEEE Conference on Computer Vision and Pattern
  Recognition (CVPR)}, 2018.

\bibitem{He2015}
Kaiming He, Xiangyu Zhang, Shaoqing Ren, and Jian Sun.
\newblock Deep residual learning for image recognition.
\newblock In {\em Proceedings of the IEEE Conference on Computer Vision and
  Pattern Recognition (CVPR)}, 2015.

\bibitem{h36m_pami}
Catalin Ionescu, Dragos Papava, Vlad Olaru, and Cristian Sminchisescu.
\newblock {Human3.6M}: Large scale datasets and predictive methods for {3D}
  human sensing in natural environments.
\newblock {\em IEEE Transactions on Pattern Analysis and Machine Intelligence
  (PAMI)}, 36(7):1325--1339, July 2014.

\bibitem{Johnson11lsp}
Sam Johnson and Mark Everingham.
\newblock Learning effective human pose estimation from inaccurate annotation.
\newblock In {\em Proceedings of Computer Vision and Pattern Recognition (CVPR)
  2011}, 2011.

\bibitem{joo2020eft}
Hanbyul Joo, Natalia Neverova, and Andrea Vedaldi.
\newblock Exemplar fine-tuning for {3D} human pose fitting towards in-the-wild
  {3D} human pose estimation.
\newblock {\em arXiv preprint arXiv:2004.03686}, 2020.

\bibitem{hmrKanazawa17}
Angjoo Kanazawa, Michael~J. Black, David~W. Jacobs, and Jitendra Malik.
\newblock End-to-end recovery of human shape and pose.
\newblock In {\em Proceedings of the IEEE Conference on Computer Vision and
  Pattern Recognition (CVPR)}, 2018.

\bibitem{kingma2014adam}
Diederik~P. Kingma and Jimmy Ba.
\newblock Adam: A method for stochastic optimization.
\newblock In {\em Proceedings of the International Conference on Learning
  Representations (ICLR)}, 2014.

\bibitem{kingma2014autoencoding}
Diederik~P Kingma and Max Welling.
\newblock Auto-encoding variational bayes, 2014.

\bibitem{Kocabas_PARE_2021}
Muhammed Kocabas, Chun-Hao~P. Huang, Otmar Hilliges, and Michael~J. Black.
\newblock {PARE}: Part attention regressor for {3D} human body estimation.
\newblock In {\em Proceedings of the International Conference on Computer
  Vision (ICCV)}, pages 11127--11137, Oct. 2021.

\bibitem{kolotouros2019spin}
Nikos Kolotouros, Georgios Pavlakos, Michael~J Black, and Kostas Daniilidis.
\newblock Learning to reconstruct {3D} human pose and shape via model-fitting
  in the loop.
\newblock In {\em Proceedings of the IEEE International Conference on Computer
  Vision (ICCV)}, 2019.

\bibitem{kolotouros2019cmr}
Nikos Kolotouros, Georgios Pavlakos, and Kostas Daniilidis.
\newblock Convolutional mesh regression for single-image human shape
  reconstruction.
\newblock In {\em Proceedings of the IEEE Conference on Computer Vision and
  Pattern Recognition (CVPR)}, 2019.

\bibitem{kolotouros2021prohmr}
Nikos Kolotouros, Georgios Pavlakos, Dinesh Jayaraman, and Kostas Daniilidis.
\newblock Probabilistic modeling for human mesh recovery.
\newblock In {\em ICCV}, 2021.

\bibitem{Lassner:UP:2017}
Christoph Lassner, Javier Romero, Martin Kiefel, Federica Bogo, Michael~J.
  Black, and Peter~V. Gehler.
\newblock {Unite the People}: Closing the loop between {3D} and {2D} human
  representations.
\newblock In {\em Proceedings of the IEEE Conference on Computer Vision and
  Pattern Recognition (CVPR)}, 2017.

\bibitem{LEE1985148}
Hsi-Jian Lee and Zen Chen.
\newblock Determination of 3d human body postures from a single view.
\newblock {\em Computer Vision, Graphics, and Image Processing},
  30(2):148--168, 1985.

\bibitem{Li_2019_CVPR}
Chen Li and Gim~Hee Lee.
\newblock Generating multiple hypotheses for {3D} human pose estimation with
  mixture density network.
\newblock In {\em Proceedings of the IEEE Conference on Computer Vision and
  Pattern Recognition (CVPR)}, June 2019.

\bibitem{li2020hybrik}
Jiefeng Li, Chao Xu, Zhicun Chen, Siyuan Bian, Lixin Yang, and Cewu Lu.
\newblock Hybrik: A hybrid analytical-neural inverse kinematics solution for 3d
  human pose and shape estimation.
\newblock In {\em Proceedings of the IEEE Conference on Computer Vision and
  Pattern Recognition (CVPR)}, 2021.

\bibitem{Lin2014MicrosoftCC}
Tsung-Yi Lin, Michael Maire, Serge~J. Belongie, James Hays, Pietro Perona, Deva
  Ramanan, Piotr Doll{\'a}r, and C.~Lawrence Zitnick.
\newblock Microsoft coco: Common objects in context.
\newblock In {\em Proceedings of the European Conference on Computer Vision
  (ECCV)}, 2014.

\bibitem{SMPL:2015}
Matthew Loper, Naureen Mahmood, Javier Romero, Gerard Pons-Moll, and Michael~J.
  Black.
\newblock {SMPL}: A skinned multi-person linear model.
\newblock In {\em ACM Transactions on Graphics (TOG) - Proceedings of ACM
  SIGGRAPH Asia}, volume~34, pages 248:1--248:16. ACM, 2015.

\bibitem{MICB19}
O. Makansi et~al.
\newblock Overcoming limitations of mixture density networks: A sampling and
  fitting framework for multimodal future prediction.
\newblock In {\em CVPR}, 2019.

\bibitem{mardia_jupp_2000}
K.~V. Mardia and P.~E. Jupp.
\newblock {\em Directional statistics}.
\newblock Wiley, 2000.

\bibitem{mohlin2020matrixfisher}
David Mohlin, Josephine Sullivan, and G\'{e}rald Bianchi.
\newblock Probabilistic orientation estimation with matrix fisher
  distributions.
\newblock In {\em Advances in Neural Information Processing Systems},
  volume~33, 2020.

\bibitem{Moon_2020_ECCV_I2L-MeshNet}
Gyeongsik Moon and Kyoung~Mu Lee.
\newblock {I2L-MeshNet}: Image-to-lixel prediction network for accurate {3D}
  human pose and mesh estimation from a single rgb image.
\newblock In {\em Proceedings of the European Conference on Computer Vision
  (ECCV)}, 2020.

\bibitem{Oikarinen2020graphmdn}
Tuomas~P. Oikarinen, Daniel~C. Hannah, and Sohrob Kazerounian.
\newblock {GraphMDN}: Leveraging graph structure and deep learning to solve
  inverse problems.
\newblock {\em CoRR}, abs/2010.13668, 2020.

\bibitem{JMLR:papamak:v22:19-1028}
George Papamakarios, Eric Nalisnick, Danilo~Jimenez Rezende, Shakir Mohamed,
  and Balaji Lakshminarayanan.
\newblock Normalizing flows for probabilistic modeling and inference.
\newblock {\em Journal of Machine Learning Research}, 22(57):1--64, 2021.

\bibitem{SMPL-X:2019}
Georgios Pavlakos, Vasileios Choutas, Nima Ghorbani, Timo Bolkart, Ahmed A.~A.
  Osman, Dimitrios Tzionas, and Michael~J. Black.
\newblock Expressive body capture: {3D} hands, face, and body from a single
  image.
\newblock In {\em Proceedings of the IEEE Conference on Computer Vision and
  Pattern Recognition (CVPR)}, 2019.

\bibitem{ravi2020pytorch3d}
Nikhila Ravi, Jeremy Reizenstein, David Novotny, Taylor Gordon, Wan-Yen Lo,
  Justin Johnson, and Georgia Gkioxari.
\newblock Accelerating {3D} deep learning with {PyTorch3D}.
\newblock {\em arXiv:2007.08501}, 2020.

\bibitem{Rezende2015normflows}
Danilo Rezende and Shakir Mohamed.
\newblock Variational inference with normalizing flows.
\newblock In Francis Bach and David Blei, editors, {\em Proceedings of the
  International Conference on Machine Learning}, volume~37 of {\em Proceedings
  of Machine Learning Research}, pages 1530--1538, Lille, France, 07--09 Jul
  2015. PMLR.

\bibitem{rupprecht2017learning}
C. Rupprecht et~al.
\newblock Learning in an uncertain world: Representing ambiguity through
  multiple hypotheses.
\newblock In {\em CVPR}, 2017.

\bibitem{saito2019pifu}
Shunsuke Saito, Zeng Huang, Ryota Natsume, Shigeo Morishima, Angjoo Kanazawa,
  and Hao Li.
\newblock Pifu: Pixel-aligned implicit function for high-resolution clothed
  human digitization.
\newblock In {\em Proceedings of the IEEE International Conference on Computer
  Vision (ICCV)}, October 2019.

\bibitem{saito2020pifuhd}
Shunsuke Saito, Tomas Simon, Jason Saragih, and Hanbyul Joo.
\newblock Pifuhd: Multi-level pixel-aligned implicit function for
  high-resolution {3D} human digitization.
\newblock In {\em Proceedings of the IEEE Conference on Computer Vision and
  Pattern Recognition (CVPR)}, June 2020.

\bibitem{STRAPS2020BMVC}
Akash Sengupta, Ignas Budvytis, and Roberto Cipolla.
\newblock Synthetic training for accurate {3D} human pose and shape estimation
  in the wild.
\newblock In {\em Proceedings of the British Machine Vision Conference (BMVC)},
  September 2020.

\bibitem{sengupta2021hierprobhuman}
Akash Sengupta, Ignas Budvytis, and Roberto Cipolla.
\newblock {Hierarchical Kinematic Probability Distributions for 3D Human Shape
  and Pose Estimation from Images in the Wild}.
\newblock In {\em Proceedings of the International Conference on Computer
  Vision}, October 2021.

\bibitem{sengupta2021probabilisticposeshape}
Akash Sengupta, Ignas Budvytis, and Roberto Cipolla.
\newblock Probabilistic {3D} human shape and pose estimation from multiple
  unconstrained images in the wild.
\newblock In {\em Proceedings of the IEEE Conference on Computer Vision and
  Pattern Recognition (CVPR)}, 2021.

\bibitem{sengupta2021semanticlocal}
Akash Sengupta, Ignas Budvytis, and Roberto Cipolla.
\newblock Probabilistic estimation of {3D} human shape and pose with a semantic
  local parametric model.
\newblock In {\em Proceedings of the British Machine Vision Conference (BMVC)},
  September 2021.

\bibitem{Sminchisescu2001covsampling}
Cristian Sminchisescu and Bill Trigg.
\newblock Covariance scaled sampling for monocular {3D} body tracking.
\newblock In {\em Proceedings of the IEEE Conference on Computer Vision and
  Pattern Recognition (CVPR)}, 2001.

\bibitem{Sminchisescu2002hyper}
Cristian Sminchisescu and Bill Trigg.
\newblock Hyperdynamics importance sampling.
\newblock In {\em Proceedings of the European Conference on Computer Vision
  (ECCV)}, 2002.

\bibitem{Sminchisescu2003kinematicjump}
Cristian Sminchisescu and Bill Trigg.
\newblock Kinematic jump processes for monocular {3D} human tracking.
\newblock In {\em Proceedings of the IEEE Conference on Computer Vision and
  Pattern Recognition (CVPR)}, 2003.

\bibitem{sun2019hrnet}
Ke Sun, Bin Xiao, Dong Liu, and Jingdong Wang.
\newblock Deep high-resolution representation learning for human pose
  estimation.
\newblock In {\em Proceedings of the IEEE Conference on Computer Vision and
  Pattern Recognition (CVPR)}, 2019.

\bibitem{tan2017}
Vince J.~K. Tan, Ignas Budvytis, and Roberto Cipolla.
\newblock Indirect deep structured learning for {3D} human shape and pose
  prediction.
\newblock In {\em Proceedings of the British Machine Vision Conference (BMVC)},
  2017.

\bibitem{varol18_bodynet}
G{\"u}l Varol, Duygu Ceylan, Bryan Russell, Jimei Yang, Ersin Yumer, Ivan
  Laptev, and Cordelia Schmid.
\newblock {BodyNet}: Volumetric inference of {3D} human body shapes.
\newblock In {\em Proceedings of the European Conference on Computer Vision
  (ECCV)}, 2018.

\bibitem{varol17_surreal}
G{\"u}l Varol, Javier Romero, Xavier Martin, Naureen Mahmood, Michael~J. Black,
  Ivan Laptev, and Cordelia Schmid.
\newblock Learning from synthetic humans.
\newblock In {\em Proceedings of the IEEE Conference on Computer Vision and
  Pattern Recognition (CVPR)}, 2017.

\bibitem{vonMarcard2018}
Timo von Marcard, Roberto Henschel, Michael Black, Bodo Rosenhahn, and Gerard
  Pons-Moll.
\newblock Recovering accurate {3D} human pose in the wild using {IMUs} and a
  moving camera.
\newblock In {\em Proceedings of the European Conference on Computer Vision
  (ECCV)}, 2018.

\bibitem{Wehrbein2021posenormflows}
Tom Wehrbein, Marco Rudolph, Bodo Rosenhahn, and Bastian Wandt.
\newblock Probabilistic monocular {3D} human pose estimation with normalizing
  flows.
\newblock In {\em Proceedings of the IEEE International Conference on Computer
  Vision (ICCV)}, 2021.

\bibitem{winkler2019learning}
Christina Winkler, Daniel Worrall, Emiel Hoogeboom, and Max Welling.
\newblock Learning likelihoods with conditional normalizing flows, 2019.

\bibitem{xu2020ghum}
Hongyi Xu, Eduard~Gabriel Bazavan, Andrei Zanfir, William~T Freeman, Rahul
  Sukthankar, and Cristian Sminchisescu.
\newblock Ghum \& ghuml: Generative 3d human shape and articulated pose models.
\newblock In {\em Proceedings of the IEEE/CVF Conference on Computer Vision and
  Pattern Recognition}, pages 6184--6193, 2020.

\bibitem{yu15lsun}
Fisher Yu, Yinda Zhang, Shuran Song, Ari Seff, and Jianxiong Xiao.
\newblock {LSUN}: Construction of a large-scale image dataset using deep
  learning with humans in the loop.
\newblock {\em arXiv preprint arXiv:1506.03365}, 2015.

\bibitem{zanfir2021hund}
Andrei Zanfir, Eduard~Gabriel Bazavan, Mihai Zanfir, William~T. Freeman, Rahul
  Sukthankar, and Cristian Sminchisescu.
\newblock Neural descent for visual {3D} human pose and shape.
\newblock In {\em Proceedings of the IEEE Conference on Computer Vision and
  Pattern Recognition (CVPR)}, 2021.

\bibitem{Zanfir_2018_CVPR}
Andrei Zanfir, Elisabeta Marinoiu, and Cristian Sminchisescu.
\newblock Monocular {3D} pose and shape estimation of multiple people in
  natural scenes - the importance of multiple scene constraints.
\newblock In {\em Proceedings of the IEEE Conference on Computer Vision and
  Pattern Recognition (CVPR)}, 2018.

\bibitem{zhang2019danet}
Hongwen Zhang, Jie Cao, Guo Lu, Wanli Ouyang, and Zhenan Sun.
\newblock Danet: Decompose-and-aggregate network for {3D} human shape and pose
  estimation.
\newblock In {\em Proceedings of the 27th ACM International Conference on
  Multimedia}, pages 935--944, 2019.

\bibitem{Zhou_2019_CVPR}
Yi Zhou, Connelly Barnes, Lu Jingwan, Yang Jimei, and Li Hao.
\newblock On the continuity of rotation representations in neural networks.
\newblock In {\em Proceedings of the IEEE Conference on Computer Vision and
  Pattern Recognition (CVPR)}, 2019.

\end{thebibliography}
}

\ifarxiv \clearpage \appendix
\label{sec:appendix}

This supplementary material contains further details regarding our probabilistic pose and shape prediction method, and presents additional quantitative and qualitative results and comparisons with other methods. Section \ref{sec:supmat_imp_details} describes the model architecture and synthetic training data. It also validates our approach for point estimate computation, and investigates our radial tanh transform for compact distribution support. Section \ref{sec:supmat_eval_details} discusses evaluation datasets and metrics, cropped evaluation dataset generation and directional variance visualisations. Finally, Section \ref{sec:supmat_experiments} presents comparisons with other methods.

\section{Implementation Details}
\label{sec:supmat_imp_details}

\subsection{Model architecture}

An overview of our model architecture is provided in Figure \ref{fig:method} of the main manuscript. Further details regarding the CNN encoder, shape/global rotation/camera MLP and per-body-part normalising flow modules are provided below.

We use a ResNet-18 CNN encoder \cite{He2015}, which takes a proxy representation input $\mathbf{X} \in \mathbb{R}^{H \times W \times C}$ and outputs a feature vector $\boldsymbol{\phi} \in \mathbb{R}^{512}$. The proxy representation consists of an edge-image and 2D keypoint heatmaps stacked along the channel dimension, with height $H = 256$, width $W = 256$ and channels $C = 18$. The choices of proxy representation and CNN encoder follow \cite{sengupta2021hierprobhuman}.

The input features $\boldsymbol{\phi}$ are passed through the shape/global rotation/camera MLP, which outputs the parameters of a Gaussian distribution over SMPL \cite{SMPL:2015} shape, $\boldsymbol{\mu}_\beta, \boldsymbol{\sigma}^2_\beta \in \mathbb{R}^{10}$, as well as deterministic estimates of weak-perspective camera parameters $\pmb{\pi} = [s, t_x, t_y] \in \mathbb{R}^3$ and the global body rotation $\mathbf{R}_\text{glob} \in SO(3)$. The latter is predicted using the continuous 6D rotation representation proposed by \cite{Zhou_2019_CVPR}, then converted to a rotation matrix. The shape/global rotation/camera MLP has 1 hidden layer with 512 nodes and ELU activation \cite{clevert2016elu}, and an output layer with 29 nodes.

\begin{table}[t]
\centering
\small
\begin{tabular}{l c}
    \hline
    \textbf{Hyperparameter} & \textbf{Value}\\
    \hline
    Shape parameter sampling mean & 0 \\
    Shape parameter sampling std. & 1.25 \\
    Cam. translation sampling mean & (0, -0.2, 2.5) m\\
    Cam. translation sampling var. & (0.05, 0.05, 0.25) m\\
    Cam. focal length & 300.0\\
    Lighting ambient intensity range & [0.4, 0.8]\\
    Lighting diffuse intensity range & [0.4, 0.8]\\
    Lighting specular intensity range & [0.0, 0.5]\\
    Bounding box scale factor range & [0.8, 1.2]\\
    \hline
    Body-part occlusion probability & 0.1 \\
    2D joints L/R swap probability & 0.1\\
    Half-image occlusion probability & 0.05\\
    Extreme crop probability & 0.1\\
    2DKP occlusion probability & 0.1\\
    2DKP noise range & [-8, 8] pixels\\
    \hline
    \end{tabular}
\caption{List of hyperparameter values associated with synthetic training data generation and augmentation. Body-part occlusion uses the 24 DensePose \cite{Guler2018DensePose} parts. Joint L/R swap is done for shoulders, elbows, wrists, hips, knees, ankles.}
\vspace{-0.7cm}
\label{tab:supmat_synthetic_data}
\end{table}

For each SMPL body-part $i \in \{1, \ldots, 23\}$, our method outputs a normalising flow distribution over the body-part's relative rotation $\mathbf{R}_i \in SO(3)$. This is conditioned on the input features $\boldsymbol\phi$, camera estimate $\pmb{\pi}$, global rotation estimate $\mathbf{R}_\text{glob}$, a shape vector sample $\boldsymbol{\beta}$ and ancestor body-part rotation samples $\{\mathbf{R}_j\}_{j\in A(i)}$, where $A(i)$ denotes the ancestors of body-part $i$ in the SMPL kinematic tree. $\boldsymbol{\beta}$ is sampled differentiably from $\mathcal{N}(\boldsymbol{\mu}_\beta(X), \boldsymbol{\sigma}^2_\beta(X))$ using the re-parameterisation trick \cite{kingma2014autoencoding}. $\{\mathbf{R}_j\}_{j\in A(i)}$ are differentiably sampled from their own respective normalising flow rotation distributions. The conditioning variables $\{\boldsymbol{\phi}, \pmb{\pi}, \mathbf{R}_\text{glob}, \boldsymbol{\beta}, \{\mathbf{R}_j\}_{j\in A(i)} \}$ are aggregated into a context vector $\mathbf{c}_i \in \mathbb{R}^{64}$ using a context generation MLP for each body-part $i$, as shown in Figure \ref{fig:method} of the main manuscript. Each context generation MLP has 1 hidden layer with 256 nodes and ELU activation \cite{clevert2016elu}, and an output layer with 64 nodes. Note: in Eqns. \ref{eqn:joint_dist} and \ref{eqn:pose_dist} in the main manuscript, we notationally replaced the conditioning variables $\boldsymbol{\phi}, \pmb{\pi}$ and $\mathbf{R}_\text{glob}$ with $\mathbf{X}$ for simplicity, because each of these are deterministically obtained as functions of $\mathbf{X}$.

\begin{figure}[t]
    \centering
    \includegraphics[width=\linewidth]{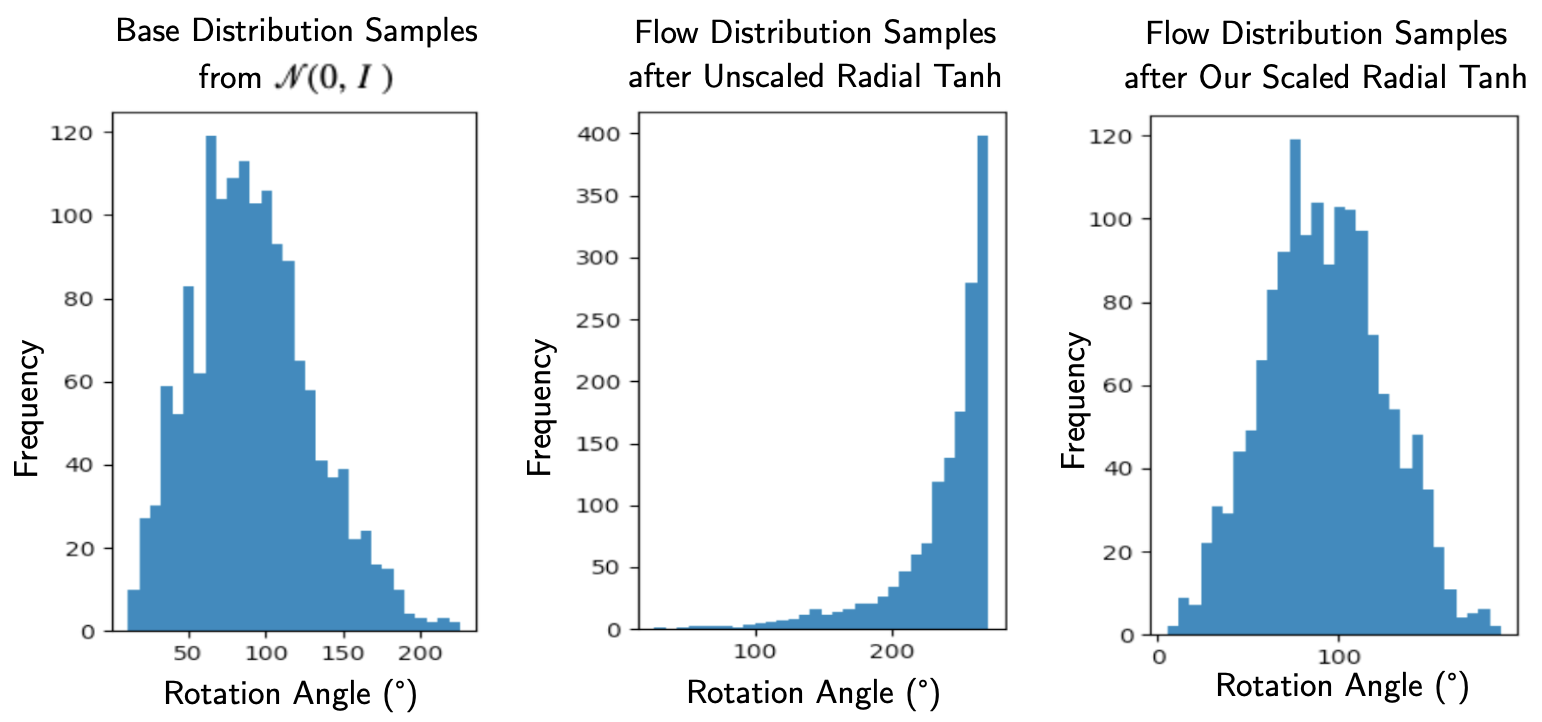}
    \caption{Comparison between the ``unscaled'' radial tanh transform proposed by \cite{falorsi2019reparameterizing} (Eqn. \ref{eqn:supmat_unscaled_radial_tanh}) and our ``scaled'' version (Eqn. \ref{eqn:supmat_radial_tanh}), in terms of sample rotation angles (or axis-angle vector magnitudes) from a randomly-initialised (i.e. un-trained) normalising flow. The unscaled transform pushes samples from the base distribution $\mathcal{N}(\mathbf{0}, \mathbf{I})$ towards the boundary of the desired support ball $B_r(\mathbf{0})$. Here, $r$ is set to $1.5\pi$ rad (i.e. 270°). This results in unstable training, as shown in Figure \ref{fig:supmat_init_flow_samples_fig}. Our scaled transform mitigates this behaviour.}
    \vspace{-0.3cm}
    \label{fig:supmat_radialtanh_fig}
\end{figure}

\begin{figure}[t]
    \centering
    \includegraphics[width=\linewidth]{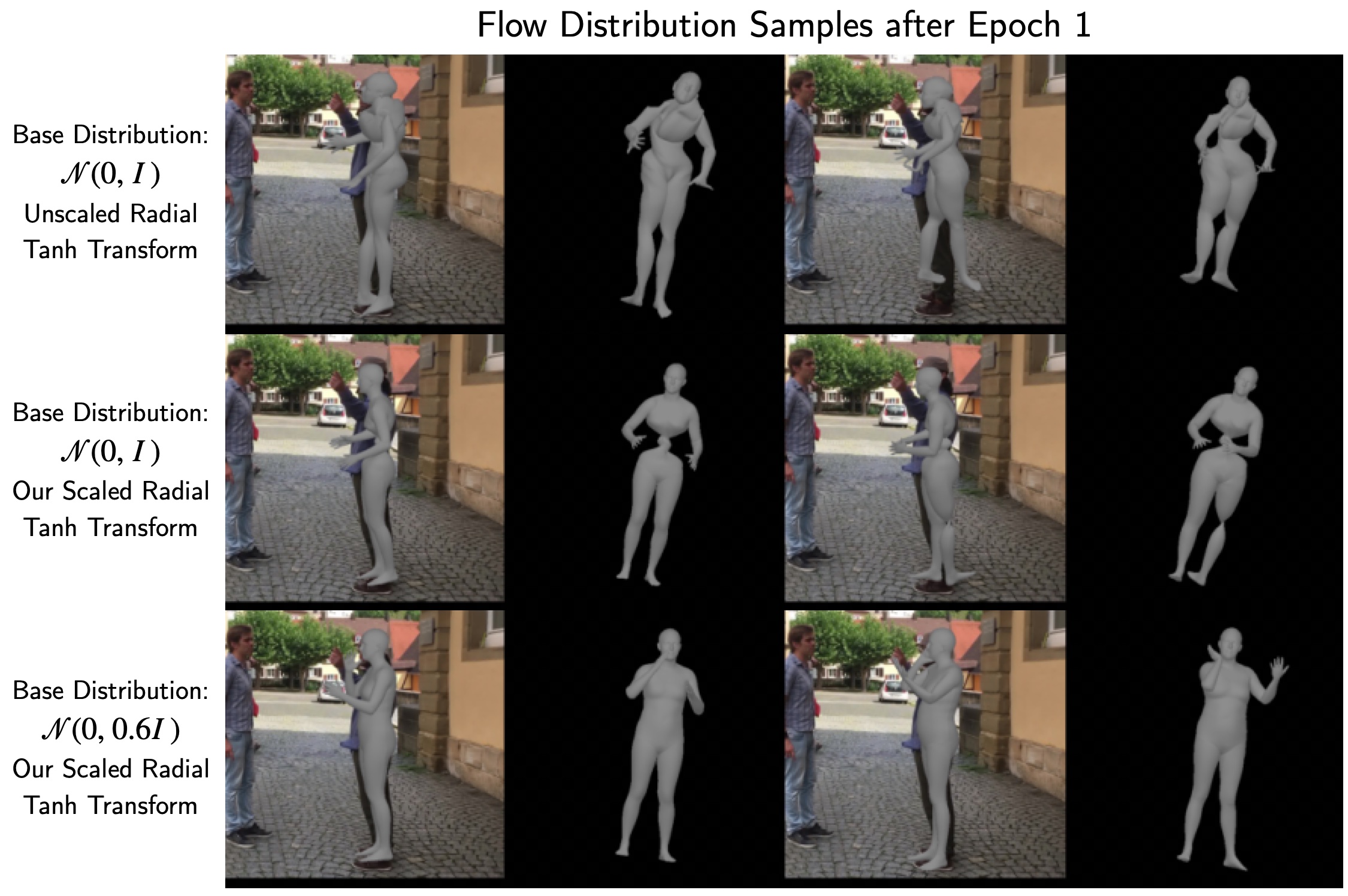}
    \caption{Effect of our ``scaled'' radial tanh transform (Eqn. \ref{eqn:supmat_radial_tanh}), and reduced-variance base distribution, on SMPL pose and shape samples after training for 1 epoch. Samples using $\mathcal{N}(\mathbf{0}, \mathbf{I})$ for base distributions and the ``unscaled'' radial tanh transform proposed by \cite{falorsi2019reparameterizing} (Eqn. \ref{eqn:supmat_unscaled_radial_tanh}) are highly unhuman (row 1). Samples become much more realistic with the scaled transform and reduced-variance base distribution (row 2 and 3). This improves training stability.}
    \vspace{-0.5cm}
    \label{fig:supmat_init_flow_samples_fig}
\end{figure}

The conditional normalising flow distribution over each body-part's rotation,  $p_{SO(3)}(\mathbf{R}_i | \mathbf{c}_i)$, is formed by pushing a flow distribution over the corresponding axis-angle vector, $p_{\mathbb{R}^3}(\mathbf{v}_i | \mathbf{c}_i)$ for $\mathbf{v}_i \in \mathbb{R}^3 \cong \mathfrak{so}(3)$, onto $SO(3)$ using the $\exp$ map (detailed by Eqn. \ref{eqn:exp_change_of_var} in the main manuscript). We use Linear Rational Spline (LRS) normalising flows \cite{dolatabadi2020lrs}, which transform a simple base distribution into a complex density function with a series of LRS coupling layer diffeomorphisms.  Specifically, let $\mathbf{z}_{k-1} \in \mathbb{R}^D$ be the input variable to the $k$-th coupling layer $f^\text{LRS}_k$, and let $\mathbf{z}_k \in \mathbb{R}^D$ be the output, such that $\mathbf{z}_k = f^\text{LRS}_k(\mathbf{z}_{k-1})$. The coupling layer \cite{dinh2015nice, dinh2017realnvp} splits the input into two parts $\mathbf{z}_{k-1}^{0:d}$ and $\mathbf{z}_{k-1}^{d:D}$. Then, the output variable is determined by
\begin{equation}
\begin{aligned}
    \mathbf{z}_k^{0:d} &= \mathbf{z}_{k-1}^{0:d}\\
    \mathbf{z}_k^{d:D} &= g(\mathbf{z}_{k-1}^{d:D}; \mathbf{w}(\mathbf{z}_{k-1}^{0:d}))
\end{aligned}
\end{equation}
where $g(\:.\:; \mathbf{w}(\mathbf{z}_{k-1}^{0:d}))$ is an \textit{element-wise} bijective and differentiable function (i.e. a diffeomorphism), whose parameters $\mathbf{w}$ depend on the first half of the input $\mathbf{z}_{k-1}^{0:d}$. Note that the Jacobian of $f^\text{LRS}_k$ is lower-triangular, and thus $\det J_{f^\text{LRS}_k}$ is easily-computed as the product of the diagonal terms of $J_{f^\text{LRS}_k}$. For an LRS coupling layer \cite{dolatabadi2020lrs}, $g$ is an element-wise spline transform. Each spline segment is a linear rational function of the form $\frac{ax+b}{cx+d}$. The parameters of $g$, $\mathbf{w}(\mathbf{z}_{k-1}^{0:d})$, are the parameters of each segment's linear rational function, and the locations of each segment's endpoints (or knots). These are obtained by passing $\mathbf{z}_{k-1}^{0:d}$ through an MLP. LRS coupling layers are able to model significantly more complex distributions \cite{dolatabadi2020lrs} than affine \cite{dinh2017realnvp} or additive \cite{dinh2015nice} coupling layers with the same number of layers composed together.

For each body-part $i$,  $p_{\mathbb{R}^3}(\mathbf{v}_i | \mathbf{c}_i)$ is implemented as an LRS-NF composed of 3 LRS coupling layer transforms, with a permutation following each coupling layer. Each layer's spline parameters are output by an MLP with 3 hidden layers that have 32 nodes each, and ELU \cite{clevert2016elu} activations.

\subsection{Radial tanh and sample angle regularisation}

We must ensure that $p_{\mathbb{R}^3}(\mathbf{v}_i | \mathbf{c}_i)$ has compact support, i.e. $p_{\mathbb{R}^3}(\mathbf{v}_i | \mathbf{c}_i) = 0$ for $\mathbf{v}_i \notin B_r(\mathbf{0})$ where $B_r(\mathbf{0})$ is an open ball of radius $\pi < r < 2\pi$. We choose $r=1.5\pi$. Towards this end, we use a radial tanh transform \cite{falorsi2019reparameterizing}, $t: \mathbb{R}^3 \rightarrow B_r(\mathbf{0})$ as the last layer of each body-part's normalising flow transform, as discussed in the main manuscript. This is reproduced here for convenience:
\begin{equation}
\label{eqn:supmat_radial_tanh}
    t(\mathbf{x}) = r \tanh\left(\frac{\|\mathbf{x}\|}{r}\right)\frac{\mathbf{x}}{\|\mathbf{x}\|}.
\end{equation}

This is slightly different to the original transform proposed in \cite{falorsi2019reparameterizing}, which is given by
\begin{equation}
\label{eqn:supmat_unscaled_radial_tanh}
    t'(x) = r \tanh\left(\|\mathbf{x}\|\right)\frac{\mathbf{x}}{\|\mathbf{x}\|} 
\end{equation}
i.e. the argument of the $\tanh$ in our transform is scaled by $1/r$. This scaling is highly beneficial for training the body pose normalising flows. To see why, consider the behaviour of $t$ and $t'$ when $\|\mathbf{x}\|$ is small, such that $\tanh\|\mathbf{x}\| \approx \|\mathbf{x}\| $. Then, $t(\mathbf{x}) \approx \mathbf{x}$ while $t'(\mathbf{x}) \approx r\mathbf{x}$. Notably, $t$ does not significantly affect points which have small magnitude, and thus are already well within the desired support $B_r(\mathbf{0})$. In contrast, $t'$ increases the magnitude of these points by a factor of $r$, pushing them towards the boundary of $B_r(\mathbf{0})$. This is illustrated by Figure \ref{fig:supmat_radialtanh_fig}, which visualises the histogram of sample magnitudes drawn from a base distribution $\mathcal{N}(\mathbf{0}, \mathbf{I})$ and the histogram of magnitudes after these samples are passed through a randomly-initialised flow ending in a radial tanh transform. Note that sample magnitudes correspond to rotation angles, as samples from $p_{\mathbb{R}^3}(\mathbf{v}_i | \mathbf{c}_i)$ are axis-angle vectors. Comparing the sample magnitude histogram for the ``unscaled'' transform \cite{falorsi2019reparameterizing} $t'$ and our ``scaled'' version $t$ demonstrates that $t'$ pushes points towards the boundary of $B_r(\mathbf{0})$ (with $r=1.5\pi$).

Large sample rotation angles from un-trained (i.e. randomly-initialised) flow distributions lead to unstable training. This is because human body-parts rarely have large rotations in natural poses, which is particularly true for torso joints in the SMPL kinematic tree. We empirically found that the flow models struggled to recover from initially too-large rotation angles during training, when using the unscaled transform $t'$. This is shown in Figure \ref{fig:supmat_init_flow_samples_fig}, where $t'$ results in extremely unhuman pose samples (row 1) after 1 training epoch, while our scaled transform $t$ gives more realistic samples. In other words, $t$ acts as a rotation angle regulariser, making use of prior domain knowledge about human body-parts often having small rotations in natural poses.

To further regularise sample rotation angles, we replace the typical base distribution $\mathcal{N}(\mathbf{0}, \mathbf{I})$ with $\mathcal{N}(\mathbf{0}, 0.6\mathbf{I})$, which has reduced variance. This results in reasonable pose samples during early training, as shown by Figure \ref{fig:supmat_init_flow_samples_fig}, row 3.

\begin{table*}[t]
\centering
\footnotesize
\renewcommand{\tabcolsep}{3pt}
\begin{tabular}{l | c c c c | c c c c | c c c c} 
\hline
\textbf{Method} & \multicolumn{4}{c|}{\textbf{3DPW}} & \multicolumn{4}{c}{\textbf{3DPW} $\mathbf{70\%}$ \textbf{Cropped}} & \multicolumn{4}{c}{\textbf{3DPW} $\mathbf{50\%}$ \textbf{Cropped}}\\
& \multicolumn{2}{c}{MPJPE (mm)} & \multicolumn{2}{c|}{MPJPE-PA (mm)} & \multicolumn{2}{c}{MPJPE (mm)} & \multicolumn{2}{c|}{MPJPE-PA (mm)} & \multicolumn{2}{c}{MPJPE (mm)} & \multicolumn{2}{c}{MPJPE-PA (mm)}\\ 
& Point & Sample Min. & Point & Sample Min. & Point & Sample Min. & Point & Sample Min.  & Point & Sample Min. & Point & Sample Min. \\
\hline
\hline
HMR \cite{hmrKanazawa17} & 130.0 & - & 76.7 & - & 177.4 & - & 96.6 & - & 214.6 & - & 120.2 & -\\
GraphCMR \cite{kolotouros2019cmr} & 119.9 & - & 70.2 & - & 120.8 & - & 74.5 & -  & 205.2 & - & 119.7 & - \\
SPIN \cite{kolotouros2019spin} & 96.9 & - & 59.0  & -  & 108.5 & - & 63.9 & - & 196.7 & - & 130.5  & -\\
PARE \cite{Kocabas_PARE_2021} & 74.5 & - & 46.5 & - & \textbf{80.0} & - & \textbf{50.6} & - & 121.8 & - & 79.7 & -\\
HybrIK \cite{li2020hybrik} & \textbf{74.1} & - & \textbf{45.0} & - & 91.9 & - & 60.7 & - & 187.5 & - & 153.0 & -\\
\hline
3D Multibodies \cite{biggs2020multibodies} & 93.8 & 74.6 \scriptsize$(\text{20.5\%})$ & 59.9 & 48.3 \scriptsize$(\text{19.4\%})$ & 110.5 & 80.9 \scriptsize$(\textbf{26.8\%})$ & 67.7 & 51.1 \scriptsize$(\text{24.5\%})$ & 190.6 & 98.4 \scriptsize$(\textbf{48.4\%})$ & 120.3 & 64.7 \scriptsize$(\textbf{46.2\%})$\\ 
Sengupta \etal \cite{sengupta2021probabilisticposeshape} & 97.1 & 84.4 \scriptsize$(
\text{13.1\%})$ & 61.1 & 52.1 \scriptsize$(\text{14.7\%})$ & 99.8 & 86.1  \scriptsize$(\text{13.7\%})$ & 62.7 & 52.2  \scriptsize$(\text{16.7\%})$ & 144.7 & 125.5 \scriptsize$(13.3\text{\%})$ & 93.6 & 76.1 \scriptsize$(\text{18.7\%})$\\
ProHMR \cite{kolotouros2021prohmr} & 97.0 & 81.5 \scriptsize$(\text{16.0\%})$ & 59.8 & 48.2 \scriptsize$(\text{19.4\%})$ & 99.4 & 84.1 \scriptsize$(\text{15.4\%})$ & 62.1 & 50.0 \scriptsize$(\text{19.5\%})$ & 143.8 & 123.3 \scriptsize$(\text{14.3\%})$ & 85.4 & 68.8 \scriptsize$(\text{19.4\%})$\\
HierProbHuman \cite{sengupta2021hierprobhuman} & 84.9 & 70.9 \scriptsize$(\text{16.5\%})$ & 53.6 & 43.8 \scriptsize$(\text{18.3\%})$ & 94.2 & 78.4 \scriptsize$(\text{16.8\%})$ & 61.6 & 49.5 \scriptsize$(\text{19.6\%})$ & 126.9 & 101.8  \scriptsize$(\text{19.8\%})$ & 87.0 & 67.7 \scriptsize$(\text{22.2\%})$\\
HuManiFlow & \textbf{83.9} & \textbf{65.1} \scriptsize$(\textbf{22.4\%})$ & \textbf{53.4} & \textbf{39.9} \scriptsize$(\textbf{25.3\%})$ & \textbf{93.5} & \textbf{71.6} \scriptsize$(\text{23.4\%})$ & \textbf{60.7} & \textbf{44.6} \scriptsize$(\textbf{26.5\%})$ & \textbf{116.4} & 86.9 \scriptsize$(\text{25.3\%})$ & \textbf{78.2} & \textbf{54.9} \scriptsize$(\text{29.8\%})$\\
\hline
\end{tabular}
\vspace{-0.1cm}
\caption{Comparison between recent deterministic (top half) and probabilistic (bottom half) pose and shape predictors in terms of accuracy on the 3DPW dataset \cite{vonMarcard2018}, as well as $50\%$ and $70\%$ cropped versions of 3DPW (see Section \ref{subsec:supmat_cropped_datasets} for cropping details). $\%$s are decreases in MPJPE(-PA) from the point-estimate to the minimum sample value computed over 100 samples. Our method, HuManiFlow, is more accurate than all current probabilistic methods. Point estimates from HuManiFlow are competitive with the state-of-the-art deterministic methods, particularly on more ambiguous and challenging cropped images.}
\label{table:supmat_3dpw_sota_comparison_accuracy}
\vspace{-0.1cm}
\end{table*}

\begin{figure}
    \centering
    \includegraphics[width=\linewidth]{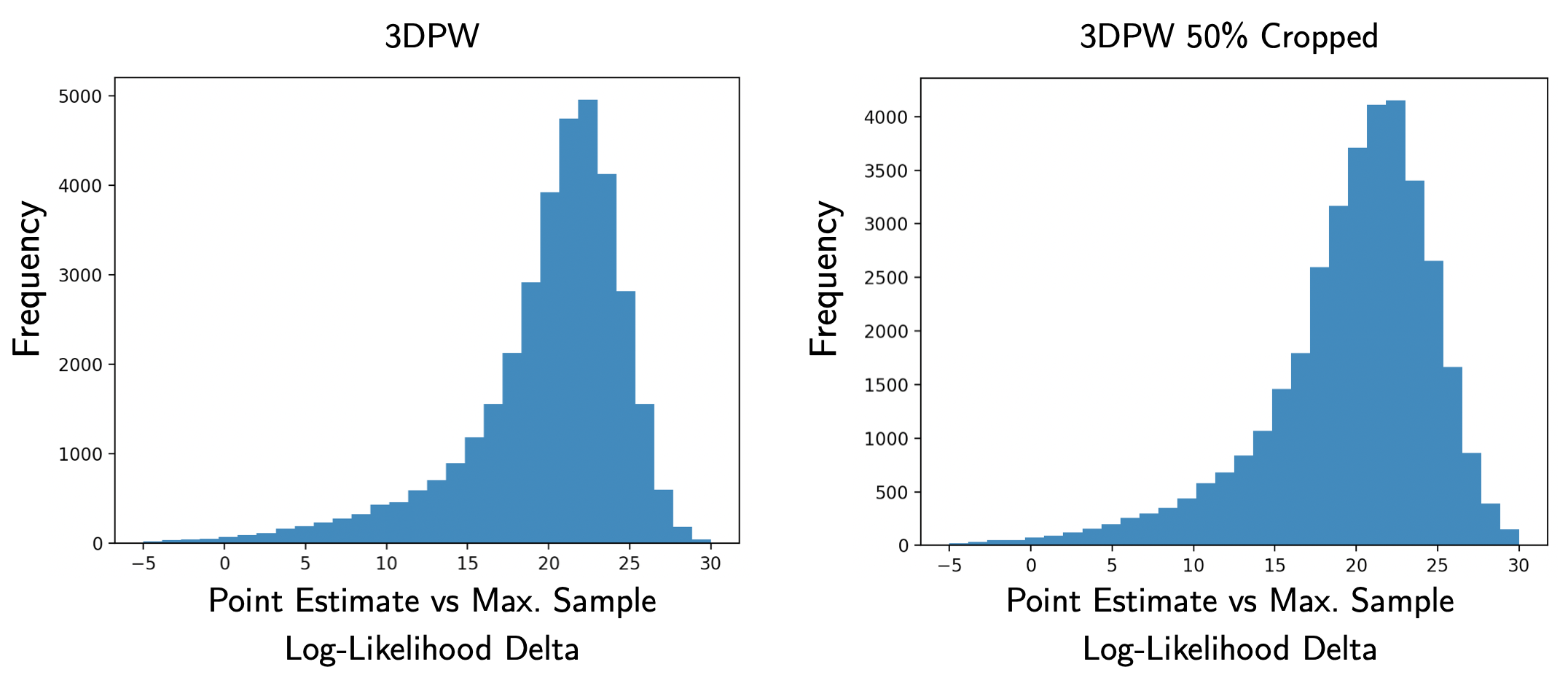}
    \caption{Histogram of differences between point estimate log-likelihoods and maximum sample log-likelihoods for 1000 samples, computed on the test set of 3DPW \cite{vonMarcard2018} and its cropped version. Point estimates generally have higher log-likelihoods than the most likely samples, as the log-likelihood differences are typically positive, confirming that the approximate method for obtaining point estimates presented in Eqn. \ref{eqn:point_estimate} of the main manuscript is suitable.}
    \label{fig:supmat_point_est_ll}
\end{figure}

\subsection{Point estimate validity}

Eqn. \ref{eqn:point_estimate} in the main manuscript describes our approach to obtaining a point estimate $(\{\mathbf{R}_i^*\}_{i=1}^{23},\boldsymbol{\beta}^*)$ from the complex joint distribution over SMPL pose and shape $p_\text{joint}(\{\mathbf{R}_i\}_{i=1}^{23}, \boldsymbol{\beta} | \mathbf{X})$ predicted by our method. The point estimate is not, in general, the actual mode of $p_\text{joint}$. However, we empirically verify that $(\{\mathbf{R}_i^*\}_{i=1}^{23},\boldsymbol{\beta}^*)$ typically has high likelihood under $p_\text{joint}$, using the test set of 3DPW \cite{vonMarcard2018}. To do so, we compute the log-likelihood of the point estimate for each test input, $p_\text{joint}(\{\mathbf{R}^*_i\}_{i=1}^{23}, \boldsymbol{\beta}^* | \mathbf{X})$, as well as the maximum sample log-likelihood $\max_{n=1,\ldots, N} p_\text{joint}(\{\mathbf{R}^n_i\}_{i=1}^{23}, \boldsymbol{\beta}^n | \mathbf{X})$ for $N=1000$ total samples. The maximum sample log-likelihood is subtracted from the point estimate log-likelihood for each test input, and the histogram of these log-likelihood deltas is visualised in Figure \ref{fig:supmat_point_est_ll}. The point estimates generally have higher log-likelihood under $p_\text{joint}$ than the most likely samples, corroborating their validity and usefulness when a single pose and shape solution is required. A qualitative comparison between point estimates from our approach and pose and shape predictions from the state-of-the-art deterministic methods \cite{Kocabas_PARE_2021, li2020hybrik} is given in Figure \ref{fig:supmat_sota_deterministic_fig}.

\subsection{Synthetic training data}

We train our pose and shape distribution prediction method using the synthetic training data generation pipeline proposed by \cite{sengupta2021hierprobhuman}. A brief overview is given below, but we refer the reader to \cite{sengupta2021hierprobhuman} for details. Moreover, hyperparameters related to synthetic data generation and augmentation are given in Table \ref{tab:supmat_synthetic_data}.

Synthetic edge-and-keypoint-heatmap proxy representations are rendered on-the-fly during training, using ground-truth SMPL \cite{SMPL:2015} pose parameters, and randomly sampled SMPL shape parameters, camera extrinsics, lighting, backgrounds and clothing. Ground-truth pose parameters are obtained from the training splits of 3DPW \cite{vonMarcard2018}, UP-3D \cite{Lassner:UP:2017} and Human3.6M \cite{h36m_pami}, giving a total of 91106 training poses, as well as 33347 validation poses from the corresponding validation splits. SMPL shape parameters are sampled from $\mathcal{N}(\boldsymbol{\beta}; \mathbf{0}, 1.25\mathbf{I})$. RGB clothing textures are obtained from SURREAL \cite{varol17_surreal} and MultiGarmentNet \cite{bhatnagar2019mgn}, which contain 917 training textures and 108 validation textures. Random backgrounds are obtained from a subset of LSUN \cite{yu15lsun} with 397582 training backgrounds and 3000 validation backgrounds. On-the-fly rendering of training inputs is done using Pytorch3D \cite{ravi2020pytorch3d}, with a perspective camera model and Phong shading. Camera and lighting parameters are randomly sampled, with hyperparameters given in Table \ref{tab:supmat_synthetic_data}.

To bridge the synthetic-to-real domain gap, synthetic proxy representations are augmented using random body-part occlusion, 2D keypoint occlusion, noise and swapping, and extreme cropping, as detailed in Table \ref{tab:supmat_synthetic_data}.

\begin{figure*}
    \centering
    \includegraphics[width=\linewidth]{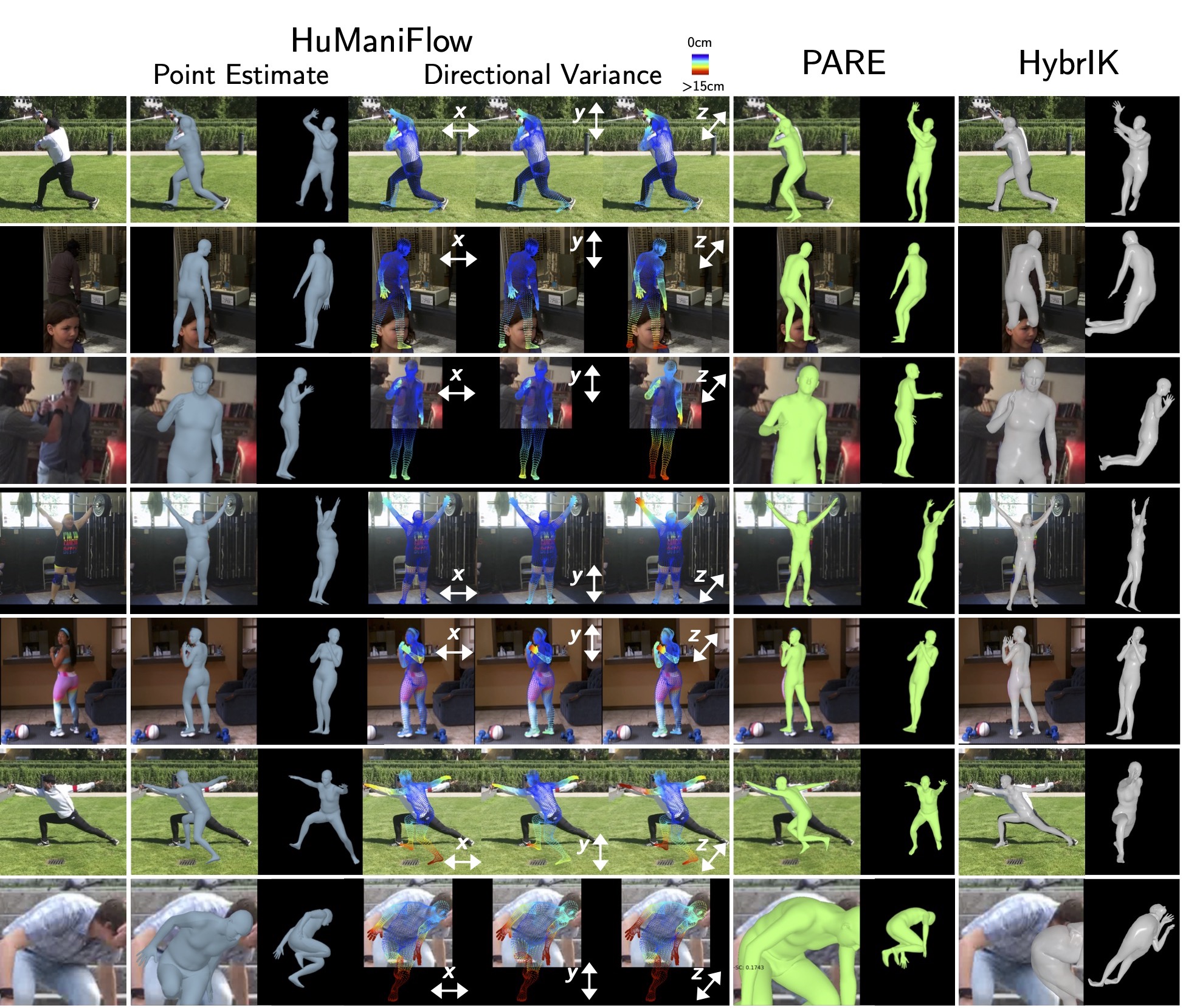}
    \vspace{-0.1cm}
    \caption{Qualitative comparison between point estimates from our probabilistic method (HuManiFlow) and the state-of-the-art single-solution (i.e. deterministic) SMPL predictors PARE \cite{Kocabas_PARE_2021} and HybrIK \cite{li2020hybrik}. HybrIK gives highly accurate solutions on less-ambiguous images, but struggles with occlusion and truncation. Point estimates from HuManiFlow and PARE perform similarly, but predicting a distribution over pose and shape allows HuManiFlow to additionally estimate prediction uncertainty, which is visualised as directional per-vertex variance. The bottom two rows show some failure cases of our method, when faced with very challenging poses or extreme truncation. The estimated uncertainty is very high for these inputs, which may be used as a signal to discount the predictions as inaccurate.}
    \label{fig:supmat_sota_deterministic_fig}
\end{figure*}

\section{Evaluation Details}
\label{sec:supmat_eval_details}

\subsection{Datasets and metrics}

The 3DPW \cite{vonMarcard2018} and SSP-3D \cite{STRAPS2020BMVC} datasets are used for our ablation studies and comparison with other human pose and shape estimation approaches. The test set of 3DPW consists of 35515 images of 2 subjects paired with ground-truth SMPL parameters and 2D keypoint locations. SSP-3D consists of 311 images of 62 subjects with diverse body shapes, paired with pseudo-ground-truth SMPL parameters and 2D keypoint locations.

As discussed in Section \ref{sec:imp_details} of the main manuscript, we use mean-per-joint-position-error (MPJPE) and MPJPE after Procrustes analysis (MPJPE-PA) to evaluate the accuracy of our method. Both MPJPE and MPJPE-PA are in units of mm. Following \cite{hmrKanazawa17, kolotouros2019spin, kolotouros2019cmr, kolotouros2021prohmr}, MPJPE and MPJPE-PA are computed using the 14 LSP joint convention \cite{Johnson11lsp}. Sample-input consistency is quantified using 2D keypoint reprojection error (or 2DKP Error) between GT \textit{visible} 2DKPs and 2DKPs computed from predicted samples, for which we use the 17 COCO keypoint convention \cite{Lin2014MicrosoftCC}. 2DKP Error is in units of pixels, assuming a $256 \times 256$ input image. Sample diversity is measured using the spread (i.e. average Euclidean distance from the mean) of 3D visible/invisible keypoints, which is denoted as 3DKP Spread. This is also computed using the 17 COCO keypoints. 3DKP Spread is in units of mm. We recognise that 3DKP Spread is flawed as a diversity metric, since the  average Euclidean distance from the mean of 3DKPs may be too simplistic to accurately reflect the diversity of 3D body pose (i.e. body-part rotation) samples, particularly when evaluating highly complex multi-modal distributions. Future work can investigate improved diversity metrics for body pose distributions.

\begin{table}[t]
\centering
\small
\renewcommand{\tabcolsep}{3.5pt}
\begin{tabular}{l l| c c } 
\hline
\textbf{Dataset} & \textbf{Method} & \textit{Consistency} & \textit{Diversity}\\
& & 2DKP Error & 3DKP Spread\\
&  & Point / Samples & Vis. / Invis.\\
\hline
\hline
\multirow{4}{0.125\linewidth}{\textbf{3DPW} $\mathbf{70\%}$ \textbf{Cropped}} & Sengupta \etal \cite{sengupta2021probabilisticposeshape} & 7.6 / 9.9  &  39.7/ 97.2 \\
& 3D Multibodies \cite{biggs2020multibodies} & 8.1 / 11.7 & 66.5 / 125.9 \\
& ProHMR \cite{kolotouros2021prohmr} & 8.1 / 9.2 & 32.0 / 60.1 \\
& HierProbHuman \cite{sengupta2021hierprobhuman} & \textbf{7.2} / 9.5 & 41.8 / 102.3 \\
& HuManiFlow & \textbf{7.2} / \textbf{8.6} & 41.9 / 116.9 \\
\hline
\multirow{4}{0.125\linewidth}{\textbf{SSP-3D} $\mathbf{70\%}$ \textbf{Cropped}} & Sengupta \etal \cite{sengupta2021probabilisticposeshape} & 9.8 / 14.3  & 60.2 / 131.6 \\
& 3D Multibodies \cite{biggs2020multibodies} & 10.5 / 15.1 & 85.1 / 160.1 \\
& ProHMR \cite{kolotouros2021prohmr} &  9.0 / 10.2 &  37.7 / 64.4\\
& HierProbHuman \cite{sengupta2021hierprobhuman} &  7.0 / 9.8  & 55.0 / 107.1 \\
& HuManiFlow & \textbf{6.9} / \textbf{8.6} & 46.9 / 123.3 \\
\hline
\end{tabular}
\caption{Comparison between probabilistic pose and shape predictors in terms of sample-input consistency and sample diversity on $70\%$ cropped versions of 3DPW \cite{vonMarcard2018} and SSP-3D \cite{STRAPS2020BMVC}. Our method, HuManiFlow, yields the most input-consistent samples (lowest visible 2DKP error) with reasonable diversity (3DKP spread).}
\label{table:supmat_3dpw_sota_comparison_diversity_consistency}
\end{table}

\subsection{Cropped dataset generation}
\label{subsec:supmat_cropped_datasets}

To evaluate our method on highly ambiguous and challenging test inputs, we generate cropped versions of 3DPW \cite{vonMarcard2018} and SSP-3D \cite{STRAPS2020BMVC}. Cropped test images are computed from the (already pre-processed) full-view test images by (i) centering at approximately the midpoint of the subject's torso, (ii) taking a square crop with dimensions given by $\alpha \%$ of the full-view image dimensions $256 \times 256$, and (iii) resizing back to $256 \times 256$. In the main manuscript, we used $\alpha = 50\%$ for all experiments with cropped data. However, the cropping percentage may be varied to evaluate our method on images with different levels of ambiguity. In Section \ref{sec:supmat_experiments} of this supplementary material, we present additional results with $\alpha = 70\%$. Examples of $50\%$ and $70\%$ cropped test images are given in Figure \ref{fig:supmat_cropped_results}.

\subsection{Directional variance visualisation}

Figures \ref{fig:intro} and \ref{fig:comparison_fig} in the main manuscript, and Figure \ref{fig:supmat_cropped_results} in this supplementary material, visualise the per-vertex directional variance of samples drawn from predicted SMPL pose and shape distributions. For a given input image, per-vertex directional variance is computed by (i) drawing $N$ samples from the predicted distribution $\{\{\mathbf{R}^n_i\}_{i=1}^{23}, \boldsymbol{\beta}^n\}_{n=1}^N$, (ii) passing each of these through the SMPL \cite{SMPL:2015} model to obtain $N$ vertex meshes $\{\mathbf{V}^n\}_{n=1}^N$ (where each $\mathbf{V}^n \in \mathbb{R}^{6890 \times 3}$) and (iii) computing the variance (more specifically, the standard deviation) of each vertex along each of the x/y/z directions, or axes. We use $N=100$. Note that the coordinate axes are aligned with the image plane, such that the x-axis represents the horizontal direction on the image, the y-axis represents the vertical direction on the image and the z-axis represents depth perpendicular to the image.

\section{Experimental Results}
\label{sec:supmat_experiments}

\nbf{Results on $70\%$ cropped images} The main manuscript reported results on the 3DPW \cite{vonMarcard2018} and SSP-3D \cite{STRAPS2020BMVC} datasets, and $50\%$ cropped versions that are more ambiguous. In Tables \ref{table:supmat_3dpw_sota_comparison_accuracy} and \ref{table:supmat_3dpw_sota_comparison_diversity_consistency} of this supplementary material, we report additional results on $70\%$ cropped versions. Comparing this with the results in the main manuscript illustrates the behaviour of pose and shape prediction methods as ambiguity increases due to greater cropping/truncation. In Table \ref{table:supmat_3dpw_sota_comparison_accuracy}, we reproduce some of the metrics previously presented in the main manuscript (on 3DPW and 3DPW $50\%$ Cropped), for convenience. Figure \ref{fig:supmat_cropped_results} presents a qualitative comparison between our method and other probabilistic pose and shape predictors using original, $70\%$ and $50\%$ cropped images.

\nbf{Qualitative comparison with deterministic methods} Figure \ref{fig:supmat_sota_deterministic_fig} compares point estimates from our method with the state-of-the-art single-solution (i.e. deterministic) monocular SMPL prediction approaches \cite{Kocabas_PARE_2021, li2020hybrik}. Figure \ref{fig:supmat_sota_deterministic_fig} also illustrates some failure cases of our method (bottom two rows) due to challenging poses and extreme truncation. We note that our approach also tends to over-estimate body shape when the subject is wearing baggy clothes, which is due to the low-fidelity synthetic training data pipeline we adopt from \cite{sengupta2021hierprobhuman}.

\nbf{Our ablation models vs. competing methods} Table \ref{table:3dpw_ablation_distributions} in the main manuscript presents our ablation study comparing several different SMPL pose distribution modelling approaches. Some of these ablation ablation models are, in fact, very similar to previously proposed probabilistic SMPL prediction methods. Specifically, the Gaussian distribution over full-body concatenated axis-angles (row 1 of Table \ref{table:3dpw_ablation_distributions}) is similar to \cite{sengupta2021probabilisticposeshape}. The normalising flow distribution over full-body concatenated axis-angles (row 3 of Table \ref{table:3dpw_ablation_distributions}) is similar to ProHMR \cite{kolotouros2021prohmr}. However, we use linear rational spline coupling layers \cite{dolatabadi2020lrs}, which are more expressive than the additive coupling layers \cite{dinh2015nice} used by \cite{kolotouros2021prohmr}. Moreover, we do not use a 6D rotation representation \cite{Zhou_2019_CVPR} for distribution prediction, to avoid the need for an orthogonality-enforcing loss, and take into account the non-Euclidean structure of $SO(3)$. Finally, the Matrix-Fisher distribution over body-part rotations (row 7 of Table \ref{table:3dpw_ablation_distributions}) is similar to HierProbHumans \cite{sengupta2021hierprobhuman}. However, as noted in the main manuscript, \cite{sengupta2021hierprobhuman} conditions body-part rotations on ancestor \textit{distribution parameters}, while we condition directly on ancestor \textit{rotations}. Our approach is more akin to a usual autoregressive model, and allows our method to be more input-consistent. For fairness, we re-train HierProbHumans with the same losses as HuManiFlow, and report results in Figure \ref{fig:supmat_rebuttal_param_vs_rot_cond}. 

\begin{figure}[t!]
    \centering
    \includegraphics[width=\columnwidth]{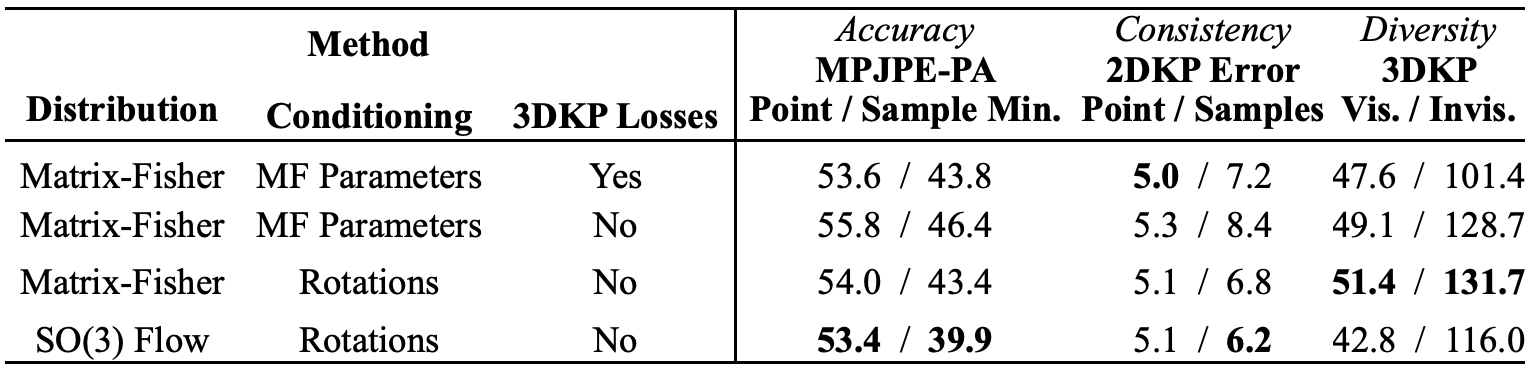}
    \caption{\textbf{Conditioning on rotation samples vs Matrix-Fisher parameters}, evaluated on 3DPW. Row 1 is HierProbHumans \cite{sengupta2021hierprobhuman}. Row 4 is HuManiFlow. Row 2 is HierProbHumans trained with the same losses as HuManiFlow - i.e. no point-based 3DKP losses. This increases diversity, but accuracy and consistency suffer. Row 3 improves these and maintains diversity, by changing HierProbHumans to \textit{condition on rotations}. This suggests that rotation-conditioning without point-based losses performs best. All models have the same backbone and no. of parameters (approx.), and are trained on the same data.}
    \vspace{-0.1in}
    \label{fig:supmat_rebuttal_param_vs_rot_cond}
\end{figure}

\begin{figure*}[h!]
    \centering
    \includegraphics[width=0.94\linewidth]{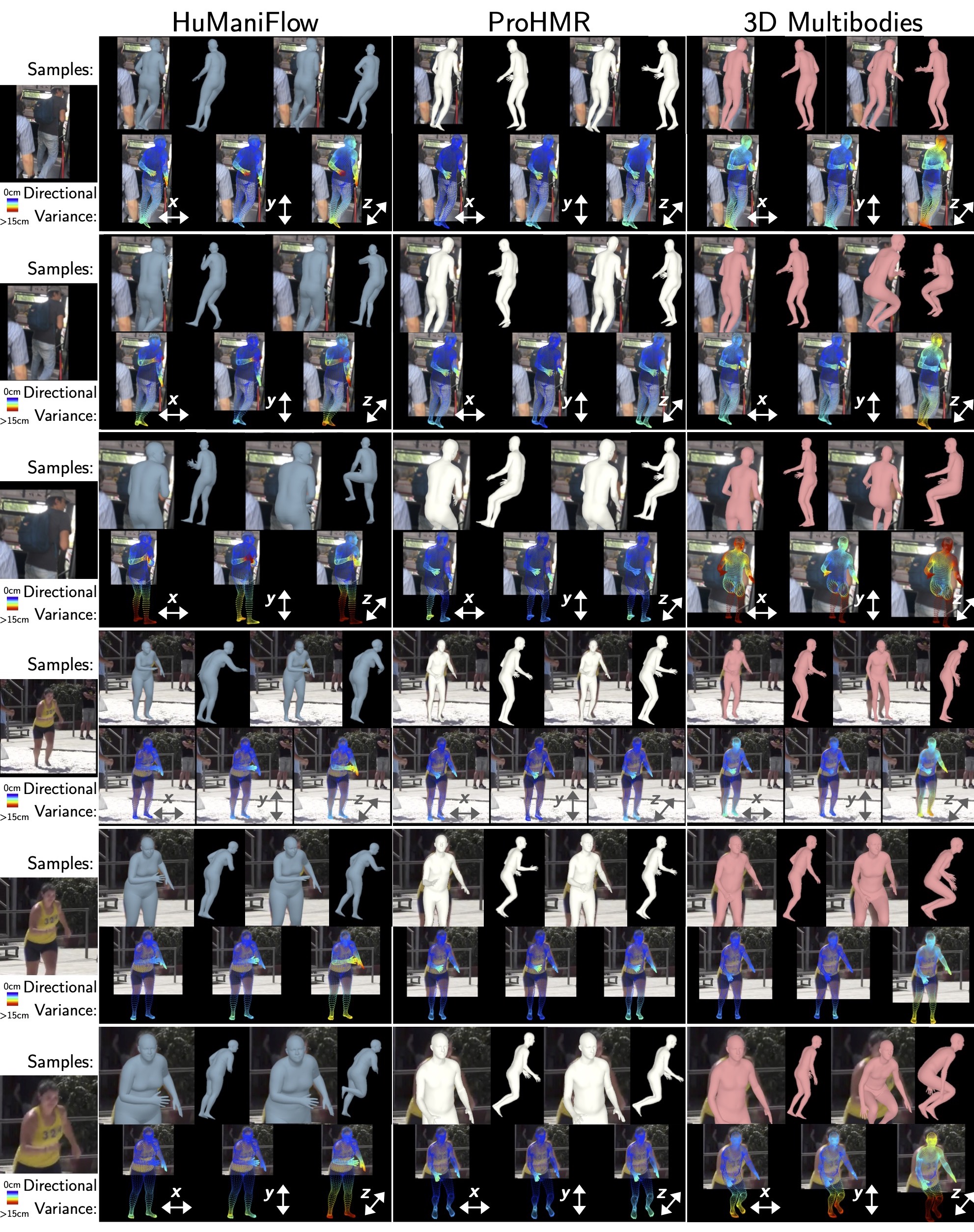}
    \vspace{-0.1cm}
    \caption{Qualitative comparison between our method (HuManiFlow), ProHMR \cite{kolotouros2021prohmr} and 3D Multibodies \cite{biggs2020multibodies} on original, $50\%$ cropped and $70\%$ cropped images (cropping details given in Section \ref{subsec:supmat_cropped_datasets}). HuManiFlow yields more \textit{diverse} pose and shape samples than ProHMR, and more \textit{input-consistent }samples than 3D Multibodies. The directional variance visualisation shows that HuManiFlow captures prediction uncertainty due to depth ambiguity (z-axis), occlusions and truncations (all-axes) in a more interpretable manner than \cite{kolotouros2021prohmr} and \cite{biggs2020multibodies}.}
    \label{fig:supmat_cropped_results}
\end{figure*} \fi

\end{document}